\documentclass{article}

\usepackage{microtype}
\usepackage{graphicx}
\usepackage{subcaption}
\usepackage{booktabs}
\usepackage{arydshln}
\usepackage{hyperref}

\usepackage[preprint]{icml2026}

\usepackage{amsmath}
\usepackage{amssymb}
\usepackage{mathtools}
\usepackage{amsthm}
\usepackage[capitalize,noabbrev]{cleveref}

\theoremstyle{plain}

\theoremstyle{definition}

\theoremstyle{remark}

\usepackage[textsize=tiny]{todonotes}
\newcommand{\rev}[1]{#1} %
\newcommand{\sd}[1]{{\color{gray}\tiny\,$\pm$\,#1}} %
\newcommand{\ci}[2]{{\color{gray}\tiny\,[#1,\,#2]}} %

\icmltitlerunning{What Do Differentiable Rule Learners Need?}

\begin{document}

\twocolumn[
  \icmltitle{What Do Differentiable Rule Learners Need? FLARE: Fuzzy Rules with Additive Evidence}

  \begin{icmlauthorlist}
    \icmlauthor{Javier Fumanal-Idocin}{essex}
    \icmlauthor{Javier Andreu-Perez}{essex}
  \end{icmlauthorlist}

  \icmlaffiliation{essex}{School of Computer Science and Electronic Engineering, University of Essex, Colchester, United Kingdom}

  \icmlcorrespondingauthor{Javier Fumanal-Idocin}{j.fumanal-idocin@essex.ac.uk}
  \icmlcorrespondingauthor{Javier Andreu-Perez}{j.andreu-perez@essex.ac.uk}

  \icmlkeywords{Explainable AI, Rule Learning, Interpretable by design}

  \vskip 0.3in
]

\printAffiliationsAndNotice{}

\begin{abstract}
	Rule learners are used where transparency and accountability are required, because they give practitioners patterns they can read. Learning rules is a discrete problem, so neurosymbolic pipelines that use rules as explanations rely on relaxations such as the straight-through estimator. However, these differentiable rule learners often lag behind both classical rule learners and black-box classifiers in terms of performance. 
	We study which parts of a rule-based classifier affect performance, and how well gradient-based training learns each of them. We found that most of the performance comes from how the rules are weighted: random rules with learned weights are already more accurate than every published differentiable rule learner we tested. Based on our results, we propose FLARE (Fuzzy Learned-Antecedent Rule Evidence), a differentiable rule learner that learns simple rules that combine to make the final decision. We also propose a learned encoding for variables with thousands of categories. We test FLARE on tabular data, using 42 datasets of OpenML-CC18 and the 38 classification tasks of TabArena. On CC18, FLARE exceeds the compared rule learners by 7.69 to 12.38 macro-F1 points and performs comparably to random forests, while remaining 1.00 to 1.11 points below tuned gradient-boosted trees. On TabArena it ranks at the same level as random forests and ahead of the other rule learners. As the concept-to-label head of a concept bottleneck model, FLARE is the best rule-based head in five of six settings. On graphs, it is competitive on PubMed and EMOGI but falls behind graph neural networks on Cora and Citeseer.
	Our code is available at \url{https://github.com/Fuminides/FLARE}.
\end{abstract}

\section{Introduction}
Explainable Artificial Intelligence (XAI) research has been motivated by the need for transparent systems in high-stakes decisions, such as finance, medicine or industrial applications \citep{rudin2019stop,arrieta2020explainable}. Interpretable models let the model explain its decision to another stakeholder in the decision, which is usually a human, but can be another judge, like an LLM \citep{zheng2023judging}.
However, due to the lagging performance of interpretable models compared to the most powerful neural baselines, it is common to use post-hoc methods to explain the decisions of opaque models, for example, SHAP or similar Shapley-based measures \citep{lundberg2017shap}. Other alternatives include generative models that produce visual explanations \citep{lang2021explaining}, the search for circuits in large language models through mechanistic interpretability \citep{olah2020zoom,conmy2023automated}, and gradient maps \citep{selvaraju2020grad}. However, these explanations can be problematic as well, as they can lack fidelity to the original decision \citep{adebayo2018sanity,slack2020fooling}, can be expensive to compute \citep{vandenbroeck2022tractability}, or might miss a global perspective of the model \citep{lundberg2020local}. Because of that, research in interpretability has been trying to obtain transparent models that can close the gap in performance with closed-box ones.

Recent literature has focused on both scalability, through gradient descent optimisation, and on joining linear and rule-based models in order to find different performance-interpretability trade-offs. Combining linear and rule-based models is usually done by using the rules as a kind of feature engineering, in which the regions defined by the rules have each a linear weight, as well as the individual features \citep{friedman2008rulefit}. SIRUS extracts rules from the nodes of a random forest and combines them into the final decision \citep{benard2021interpretable}. EBM learns an additive model with pairwise interactions, fitting one shallow tree per feature or pair of features \citep{lou2013accurate,nori2019interpretml}, and GRANDE trains an ensemble of hard, axis-aligned decision trees end to end by gradient descent, with instance-wise weights for the trees \citep{marton2024b}. Gradient-based rule systems have developed different variants of the straight-through estimator (STE): RRL uses gradient grafting, in which the discrete and the real models work in parallel to generate a valid gradient estimation for the backward pass while keeping a discrete forward \citep{wang2021,wang2024b}. FRR also trains its rules by gradient descent, keeping fuzzy partitions with a linguistic meaning and deciding each sample by a single rule, aiming for very small rule bases \citep{fumanal2025compact}. NeuRules learns the thresholds, the conjunctions and the order of a rule list together, each relaxed to a smooth version that is annealed towards a crisp one during training \citep{xu2025}. RL-Net also learns an ordered rule list, as a neural network whose rule neurons are DR-Net's conjunctions and whose fixed hierarchy layer lets a rule fire only when no earlier rule has \citep{dierckx2023,qiao2021}. Outside gradient-based learning, SamRuLe scales rule-list learning by searching for the rule list on a sample of the data, with guarantees on how close it is to the one learned on all of it \citep{pellegrina2024scalable}. Concept bottleneck models predict a set of concepts from which the final prediction is made, usually with a linear model \citep{koh2020concept}, although rule-based heads have also been explored \citep{ciravegna2023,barbiero2023,vemuri2026logiccbms}. However, in general, these methods tend to create very dense rule bases, and when benchmarked their performance is often not significantly better than that of classical methods \citep{fumanal2026evidential}. So, it is not clear which of the components of these learners: the conditions, the weights of the rules, their number and length, or the gradient estimator; are the ones that matter for performance.

In this work:
\begin{itemize}
	\item We empirically study the different components learned by differentiable rule learners: how they learn conditions, how they weight the rules, their length, and how the different gradient techniques work. Most surprisingly, we found that much of the performance actually comes from rule weighting: fixing random rules and then learning only their weights is more accurate than every published differentiable rule learner we tested.
	\item We propose FLARE (Fuzzy Learned-Antecedent Rule Evidence): a differentiable rule learner designed around these findings, which combines gradient-based condition search with additive rule weights, plus a continuous encoding scheme for categorical variables that supports thousands of categories. FLARE scales to tasks with 100,000 training rows, 1,776 features or more than 15,000 categories in a single variable, and to graphs with over 11,000 variables per node.
	\item We evaluate FLARE on 42 OpenML-CC18 datasets and the 38 classification tasks of TabArena. We benchmark it against seven rule learners, two differentiable tree learners and three gradient-boosted tree libraries, as well as other reference classifiers, all tuned with the same budget on CC18. Beyond tabular data, we use it as the head of concept bottleneck models and for node classification on graphs.
\end{itemize}

\section{Related Work}

\subsection{Classical Rule Learning} Rule learning has classically been done with greedy algorithms such as CART and C4.5, which grow decision trees by choosing the split that maximises information gain, or RIPPER by sequential covering \citep{breiman1984cart,quinlan1993c45,cohen1995ripper}. However, greedy search is a limited heuristic in terms of performance, which is why other alternatives have been developed as well. For example, genetic search for fuzzy rule bases \citep{alcalafdez2011fuzzy,fumanal2024ex, shiraishi2026class}. Bayesian rule lists and rule sets sample or rule structures potential rulebases \citep{letham2015,yang2017b}, and CLASSY selects rule lists by the minimum description length principle \citep{proenca2020interpretable}. Other methods solve the problem exactly or with bounds: CORELS finds certifiably optimal rule lists by branch-and-bound \citep{angelino2018}, mixed-integer programming and branch-and-bound give optimal and optimal sparse decision trees \citep{bertsimas2017,hu2019}. Another relevant branch is submodular optimisation \citep{lakkaraju2016,yang2021}. Moreover, once the optimum can be computed, greedy heuristics often turn out to be optimal or close to it \citep{hu2019}, which suggests that better search algorithms will still have a hard time consistently beating the basic greedy baselines \citep{fumanal2026evidential}.

\subsection{Gradient-based Rule Learning} The popularity of gradient-based learning has motivated the development of several approaches to turn the rule search problem into a continuous one that can be solved using backpropagation. Initial works started by using the STE to binarise the rule neurons \citep{qiao2021}, an approach later extended to learn the thresholds of continuous features \citep{zhang2023} and the splits of axis-aligned trees \citep{marton2024a}. Since then, other works have developed more sophisticated gradient techniques, like gradient grafting, in which a continuous and a discrete model run in parallel and the gradient of the loss at the discrete output is passed back through the continuous one \citep{wang2021,wang2024b}. Annealing approaches use a smooth model that is driven towards a crisp one during training \citep{xu2025, fumanal2025compact}. Other approaches have adopted fuzzy logic, in which the degree of truth is a continuous value between 0 and 1, thus naturally allowing for continuous optimisation \citep{vankrieken2022,fumanal2025compact}.
 Finally, there are also approximation-only approaches, in which the resulting rulebase only approximates the original model \citep{silva2020}. Compared to classical rule learning, these methods have also been proposed in neuro-symbolic pipelines, such as the concept-to-label head of concept bottleneck models \citep{ciravegna2023,barbiero2023,vemuri2026logiccbms} or the learning of logical rules over knowledge bases \citep{yang2017a,evans2018}.

\section{Fuzzy Learned-Antecedent Rule Evidence}
\label{sec:flare}

In this section we describe the components of FLARE. The rationale behind FLARE is based on our empirical results regarding what really works in rule-based learning in general, and in gradient-based rule learning in particular. These results are summarised in Section~\ref{sec:results_study}, and shown in full in Appendix \ref{apx:empirical_component_ablation}. For the reader convenience, a notation summary is also introduced in Appendix~\ref{apx:notation}.

\begin{figure*}[t]
	\centering
	\includegraphics[width=\textwidth]{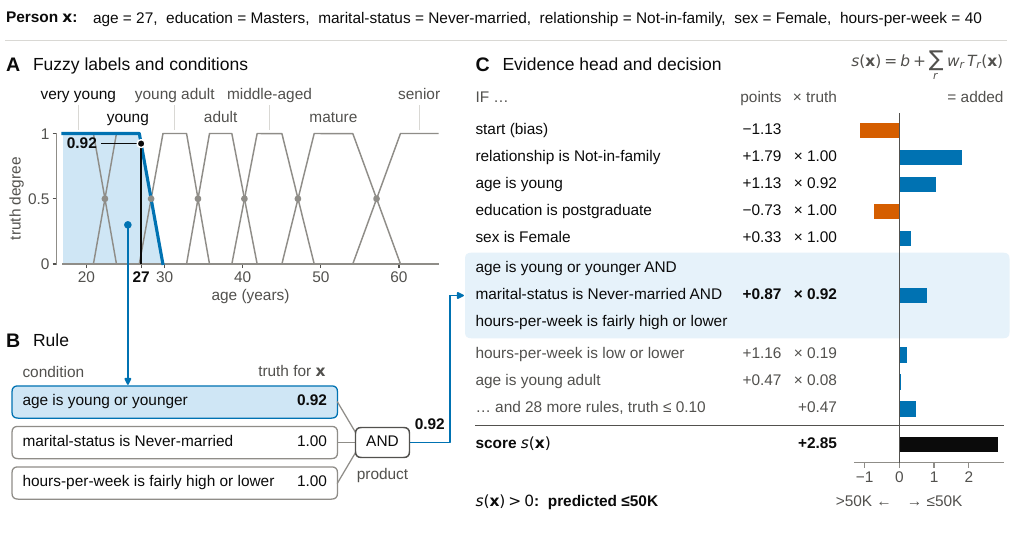}
	\caption{\textbf{FLARE scheme.} For one test person of the adult dataset. (A) Each real variable is described by fuzzy labels. (B) The truth of a rule is the product of the truths of its conditions. (C) The score starts at the bias and every rule adds its points times its truth degree, and the final decision is taken.}
	\label{fig:overview}
\end{figure*}

FLARE is a set of rules that additively scores each class (Figure~\ref{fig:overview}). Each variable is described by a few fuzzy labels, and a rule is a conjunction of up to three conditions (using the product T-norm). Let $T_r(\mathbf{x}) \in [0,1]$ be the degree to which rule $r$ holds for the sample $\mathbf{x}$. The score of class $c$ is
\begin{equation}
	s_c(\mathbf{x}) = b_c + \sum_{r=1}^{R} w_{rc}\, T_r(\mathbf{x}),
	\label{eq:score}
\end{equation}
where $b_c$ is a bias and $w_{rc}$ are the signed points that rule $r$ gives to class $c$, and the predicted class is the one with the highest score.

\subsection{Partitions in Continuous Variables}
\label{sec:partitions}
Each real variable $x_j$ is covered by $L$ trapezoidal fuzzy sets $\mu_{j1}, \dots, \mu_{jL}$ that form a Ruspini partition, which means that $\sum_{l} \mu_{jl}(x_j) = 1$ for every value. So, the partition is determined by $L$ plateaus (where the fuzzy membership is equal to 1) and $L-1$ ramps, each shared by two neighbouring labels, where no label is equal to 1. Each label is a trapezoid defined by consecutive breakpoints $p_{j,0} \le p_{j,1} \le \dots \le p_{j,2L-1}$:
\begin{equation}
	\mu_{jl}(x_j) =
	\begin{cases}
		\dfrac{x_j - p_{2l-3}}{p_{2l-2} - p_{2l-3}} & p_{2l-3} \le x_j < p_{2l-2},\\[4pt]
		1 & p_{2l-2} \le x_j \le p_{2l-1},\\[4pt]
		\dfrac{p_{2l} - x_j}{p_{2l} - p_{2l-1}} & p_{2l-1} < x_j \le p_{2l},\\[4pt]
		0 & \text{otherwise},
	\end{cases}
	\label{eq:trapezoid}
\end{equation}
where we drop the index $j$ of the breakpoints, and the first label has no rising ramp and the last one no falling ramp. Label $l$ falls on the same ramp $[p_{j,2l-1}, p_{j,2l}]$ on which label $l+1$ rises. We start from a partition in which every plateau and ramp hold the same share of the training rows: $p_{j,i}$ is the $i/(2L-1)$ quantile of $x_j$ in the training data.

Then, we can learn the lengths $\delta_{j,i} = p_{j,i} - p_{j,i-1}$ of the $2L-1$ segments, as:
\begin{equation}
	\boldsymbol{\delta}_j = (M_j - m_j)\, \mathrm{softmax}(\boldsymbol{\theta}_j),
\end{equation}
where $m_j = p_{j,0}$ and $M_j = p_{j,2L-1}$ are the smallest and largest training values of $x_j$, and $\boldsymbol{\theta}_j \in \mathbb{R}^{2L-1}$ is initialised at the logarithm of the initial lengths. This still results in partitions that always cover the whole range and whose segments never collapse to zero, which would leave a label without gradient. We call \emph{crossing} the point where two neighbouring labels are both 0.5 true, in the middle of their shared space. Training the gaps moves the crossings, and with them the cut points of every rule that uses them. There is also another reason to ensure Ruspini partitions. Because the labels sum to one, conditions such as ``at most label $k$'' ($\mu_{j1} + \dots + \mu_{jk}$) or ``at least label $k$'', can be represented as a single trapezoid that goes from 1 to 0 along one ramp.

Interpreting each fuzzy partition is also intuitive. Each label can be named after its position in the partition (for example, for $L=3$, \emph{low}, \emph{medium}, \emph{high}), or with a domain name when one is available.

\subsection{Handling Categorical Variables}
\label{sec:categorical}
A categorical variable with a small number of categories simply gets one label per category. During training, though, the other categories keep a small truth of 0.1 instead of 0 in the backward computation only, so that conditions keep receiving gradient. However, this scheme does not scale to variables with thousands of categories (most categories are too rare to support a rule of their own, and the model would risk not fitting in GPU memory). For variables with more than 32 categories in the training data, we therefore learn groups of categories. For that, we propose the following.
Each category $a$ gets a point $\mathbf{z}_a \in \mathbb{R}^{K}$, with $K$ the dimension of the space, and the $L$ labels of the variable are Gaussian regions of that space, shared by all the rules, with centers $\boldsymbol{\mu}_l$ and widths $\sigma_l$. In this space we apply a reconstruction loss with a decoder that reconstructs each category from its point:
\begin{equation}
	\mathcal{L}_{\mathrm{rec}} = -\sum_{a} \pi_a \log g_a(\mathbf{z}_a),
	\label{eq:rec_loss}
\end{equation}
where $\pi_a$ is the percentage of the training rows that have category $a$, and $g_a(\mathbf{z})$ is the probability that the decoder $\mathbf{g}$ gives to category $a$ from the point $\mathbf{z}$. The decoder is a network with one hidden layer. This loss keeps categories together if the rules gain performance from treating them similarly. Then, a radius loss pulls $\lVert \mathbf{z}_a \rVert$ towards a target that grows as $1/\sqrt{n_a}$, (being $n_a$ the number of samples of category $a$ in the training set) placing rare categories farther from the centre:
\begin{equation}
	\mathcal{L}_{\mathrm{rad}} = \sum_{a} \pi_a \left(\lVert \mathbf{z}_a \rVert - r_a\right)^2,
	\label{eq:rad_loss}
\end{equation}
where the target radius $r_a$ maps $u_a = n_a^{-1/2}$ linearly onto a fixed range $[r_{\min}, r_{\max}]$:
\begin{equation}
	r_a = r_{\min} + (r_{\max} - r_{\min})\, \frac{u_a - \min_b u_b}{\max_b u_b - \min_b u_b}.
	\label{eq:target_radius}.
\end{equation} 

The first axis of this space starts at the smoothed effect of each category on the class, as in the ordering of categories used for tree splits \citep{breiman1984cart}. 

In order to read a rule, we get the Gaussian blob in the latent space and see which categories lies in the region, this results in conditions like e.g.\ ``$x_j$ is in $\{A, F, G\}$''.

\subsection{Learning Conditions and Rules}
\label{sec:rules}
A rule has up to a fixed number $S$ of condition slots. For each slot, the rule keeps a selector over the variables and, for each variable, a selector over its labels, which all the slots of the rule share. The slot takes the variable and the label with the largest selector weights:
\begin{equation}
	j_{rs} = \arg\max_{j}\, \alpha_{rsj}, \qquad l_{rj} = \arg\max_{l}\, \beta_{rjl},
	\label{eq:selection}
\end{equation}
where $\alpha_{rsj}$ is the weight that slot $s$ of rule $r$ gives to variable $j$, and $\beta_{rjl}$ is the weight that rule $r$ gives to label $l$ of variable $j$. For a real variable, $l$ also runs over the ``at most'' and ``at least'' sets of Section~\ref{sec:partitions}. Slot $s$ of rule $r$ thus states the condition ``$x_j$ is $l_{rj}$'' for $j = j_{rs}$, and its truth is
\begin{equation}
	\tau_{rs}(\mathbf{x}) = \mu_{j\, l_{rj}}(x_j), \qquad j = j_{rs}.
	\label{eq:slot_truth}
\end{equation}
Also, every slot except the first one has a selector that can switch it off, so we can use rules that are as short as possible: for each variable $j$ the slot keeps two weights, $\gamma^{\mathrm{on}}_{rsj}$ and $\gamma^{\mathrm{off}}_{rsj}$, and it is off when $\gamma^{\mathrm{off}}_{rsj} > \gamma^{\mathrm{on}}_{rsj}$ for the variable $j = j_{rs}$ that it has selected. A slot that would repeat a condition of the same rule is also switched off to avoid duplicate conditions. Let $\mathcal{A}_r$ be the slots of rule $r$ that remain active, the truth of a rule is the product of the truths of its conditions:
\begin{equation}
	T_r(\mathbf{x}) = \prod_{s \in \mathcal{A}_r} \tau_{rs}(\mathbf{x}).
	\label{eq:rule_truth}
\end{equation}

FLARE uses a fixed mix of rules. We fix most of them to have a single condition and act as the main effects of an additive model. A smaller group uses up to three conditions and captures interactions.

\subsection{Final Decision}
\label{sec:head}
Every rule has a vector of signed points $\mathbf{w}_r \in \mathbb{R}^{C}$, one per class, and every class has a bias $b_c$ (Eq.~\ref{eq:score}). A rule that adds its points to the class scores, weighted by its truth degree.
	
\paragraph{Reducing variance} Finally, in order to reduce the variance that is typical of rule learners, FLARE trains $B$ models, each with its own split of the training rows into training and validation rows, and averages them. The first model learns the partitions, and the other $B-1$ are then trained in parallel on those partitions, so the averaged model has a single partition per variable. This, and the fact that the head is linear, makes the average again another FLARE model of the form of Eq.~\eqref{eq:score}. This does not change the nature of the rule-based reasoning, at the cost of a rule base up to $B$ times larger.

\subsection{Overall Aim and Training}
\label{sec:training}
All the parameters are trained together by gradient descent. The objective is the cross-entropy of the class scores, with class weights inversely proportional to the class frequencies, which is a standard technique for imbalanced problems:
\begin{equation}
	\mathcal{L}_{\mathrm{CE}} = -\frac{\sum_{i} \omega_{y_i} \log p_{y_i}(\mathbf{x}_i)}{\sum_{i} \omega_{y_i}},
	\label{eq:ce_loss}
\end{equation}
where $i$ runs over the rows of a batch, $y_i$ is the class of row $i$, $\omega_c$ is the weight of class $c$ and $p_c(\mathbf{x})$ is the softmax of the scores of Eq.~\eqref{eq:score}, plus the reconstruction and radius terms of the category spaces when there are any (Eqs.~\ref{eq:rec_loss} and~\ref{eq:rad_loss}):
\begin{equation}
	\mathcal{L} = \mathcal{L}_{\mathrm{CE}} + \sum_{j \in \mathcal{J}} \left(\lambda\, \mathcal{L}_{\mathrm{rec}, j} + \mathcal{L}_{\mathrm{rad}, j}\right),
	\label{eq:objective}
\end{equation}
where $\mathcal{J}$ is the set of variables with a category space and $\lambda$ weighs the reconstruction against the radius term.

The selectors make discrete choices, as each slot uses one variable and one label. We train them with gradient grafting \citep{wang2021}. The forward pass computes the discrete model (which is the one deployed), with every slot taking one single variable and label. The backward pass takes its gradient from a continuous surrogate, in which each slot is a weighted mixture of its candidates, blended with a uniform share $k$ that is annealed from 1 towards 0 during training. Every selector works in the same way. Let $\phi_1, \dots, \phi_Q$ be the weights of a selector over $Q$ candidates, whose truths for the sample are $h_1, \dots, h_Q$, and let $q^* = \arg\max_q \phi_q$ be the chosen one. The forward pass outputs $h_{q^*}$, and the backward pass uses the gradient of
\begin{equation}
	\tilde{h} = (1-k)\, \rho_{q^*} h_{q^*} + \frac{k}{Q} \sum_{q=1}^{Q} \rho_q h_q, \quad \boldsymbol{\rho} = \mathrm{softmax}(\boldsymbol{\phi} / \eta),
	\label{eq:surrogate}
\end{equation}
where $\eta$ is a fixed temperature. Besides the selectors, the backward pass also has a residual term. The forward pass uses $\log T_r$, and the backward pass uses the gradient of
\begin{equation}
	\tilde{\ell}_r = \log T_r + \min(0.1, k) \sum_{j} v_{rj}, \quad v_{rj} = \mu_{j\, l_{rj}}(x_j),
	\label{eq:residual}
\end{equation}
where $v_{rj}$ is the truth of the condition of rule $r$ on variable $j$. This term gives the condition on every variable a direct path to the truth of the rule. A concrete analysis of the equations involved in the differentiation of FLARE is detailed in Appendix \ref{apx:differentiation_specs}.
Besides this, each parameter set also learns at different speeds (Appendix~\ref{apx:experimental_details}).

\section{Experimental Settings}
\label{sec:settings}

 \subsection{Datasets and Evaluation}
\label{sec:datasets}
\paragraph{Tabular data.} We use two collections of datasets. From TabArena \citep{erickson2025tabarena} we experiment on 30 binary and 8 multi-class tasks up to 100,000 training rows, 1,776 variables and 15,416 categories in a single variable. For evaluation, we follow TabArena's own protocol, which uses its predefined train/test splits, ROC AUC for the binary tasks and log-loss for the multi-class ones, and then compares the methods by their mean rank over the tasks. We also take 42 tabular datasets from OpenML-CC18 \citep{bischl2021openml}, which have from 500 to 67,557 rows, from 4 to 118 variables and from 2 to 11 classes. For each of them we run a stratified 5-fold cross-validation with three different seeds, and we report the macro-F1 on the test rows.

\paragraph{Image data.} We test FLARE within a concept bottleneck model framework, in which a set of human-defined concepts is extracted from the image and then the class is predicted from those \citep{koh2020concept}. FLARE replaces the linear model that is usually used for the second step. The concepts are predicted by an Inception v3 \citep{szegedy2016rethinking} following \citet{koh2020concept}, and the head is trained either on the ground-truth concepts or on the ones predicted by the detector in a sequential training scheme. We use CUB with its per-image annotations (112 concepts, 200 classes) \citep{wah2011cub}, SUN Attribute (102 concepts, 717 classes) \citep{patterson2012sun} and Derm7pt (12 concepts taken from the 7-point checklist, 5 diagnosis groups) \citep{kawahara2019seven}. We report the macro-F1 obtained with the concepts predicted for the test images.

\paragraph{Graph data.} We use Cora, Citeseer and PubMed with the public split of \citet{yang2016revisiting}, where we report the accuracy, and the protein interaction network of EMOGI \citep{schultesasse2021emogi}, where the task is to identify cancer driver genes and we report the area under the precision-recall curve on the authors' 746 test genes. In graph data, each rule contains variables from a node and its neighbours. %

\paragraph{Statistical tests.} We use two-sided Wilcoxon signed-rank tests on paired dataset or task scores, averaged over splits and seeds. The tests use macro-F1 on CC18 and prediction error on TabArena: one minus ROC AUC for binary tasks and log-loss for multi-class tasks.

\subsection{Compared Methods}
\label{sec:baselines}

\paragraph{Rule and tree learners.} We compare FLARE with classical rule learners based on discrete search (RIPPER \citep{cohen1995ripper}, CORELS \citep{angelino2018} and CLASSY \citep{proenca2020interpretable}), with differentiable rule learners (RRL \citep{wang2021}, RL-Net \citep{dierckx2023}, NeuRules \citep{xu2025} and DR-Net \citep{qiao2021}) and with the differentiable tree learner GradTree \citep{marton2024a}. As additional references we also report FIGS \citep{tan2022figs}, RuleFit \citep{friedman2008rulefit} and CART \citep{breiman1984cart}. For image data, we add two rule-based heads proposed for concept bottleneck models: LogicCBM \citep{vemuri2026logiccbms} and Deep Concept Reasoner \citep{barbiero2023}.

\paragraph{Standard classifiers.} For tabular data we use gradient-boosted trees, which are the usual reference in terms of performance: XGBoost \citep{chen2016xgboost}, LightGBM \citep{ke2017lightgbm} and CatBoost \citep{prokhorenkova2018catboost}. We also report logistic regression, random forests \citep{breiman2001random}, GRANDE \citep{marton2024b}, which is an ensemble of differentiable trees, and EBM \citep{nori2019interpretml}, and in TabArena the published results of its own methods, which include neural networks and a tabular foundation model. For image data we use logistic regression, which is the standard head of a concept bottleneck model. For graph data we use SGC \citep{wu2019sgc}, GCN \citep{kipf2017gcn} and GAT \citep{velickovic2018gat}, and also logistic regression, random forests, GRANDE, EBM and the rule and tree learners above.

\paragraph{FLARE configurations.} We study two configurations of FLARE. In the first one each real variable has $L = 7$ labels, and we average $B = 8$ models of $R = 120$ rules each, of which 100 have a single condition and the other 20 can have up to $S = 3$. The second one, FLARE-\emph{explainability}, aims for a small rule base: it is a single model ($B = 1$) with $R = 40$ rules, of which 30 have a single condition.

Implementation details and hyperparameter tuning of each method are given in Appendix~\ref{apx:experimental_details}. An example of the rules of FLARE is in Appendix~\ref{apx:expanded_results}.

\section{Results}
\label{sec:results}

\subsection{Tabular Data}
\label{sec:results_tabular}

\begin{table*}[t]
\centering
\small
\caption{\textbf{Results on the two tabular benchmarks.} Score means use all completed tasks. Bold indicates the best rule or tree learner (including FLARE), and underlining indicates the best method overall in each column. $^\ddagger$Class-only output; for ranking, log-loss is computed from one-hot predictions. $^\S$Results published by TabArena. On CC18, $\pm$ is the standard deviation over the three seeds of the mean over the datasets. On TabArena, the brackets give the 95\% bootstrap interval of the mean rank over the 38 tasks (the paired tests are in Appendix~\ref{apx:tabular_paired}).}
\label{tab:tabular}
\begin{tabular}{@{}llccccc@{}}
\toprule
 & & \multicolumn{2}{c}{CC18} & \multicolumn{3}{c}{TabArena} \\
\cmidrule(lr){3-4} \cmidrule(l){5-7}
 & Method & Macro-F1 & Rank & ROC AUC & Log-loss & Rank \\
\midrule
 & FLARE & \textbf{77.67}\sd{0.11} & \textbf{5.29} & \textbf{0.825} & \textbf{0.348} & \textbf{6.63}\ci{6.18}{7.11} \\
 & \hspace{0.75em}-explainability & 74.00\sd{0.11} & 10.48 & 0.794 & 0.379 & 9.79\ci{9.26}{10.29} \\
\midrule
Rule and tree & RuleFit & 74.40\sd{0.21} & 9.79 & 0.817 & 0.461 & 7.97\ci{7.26}{8.66} \\
 baselines & CART & 73.48\sd{0.21} & 10.74 & 0.668 & 4.215 & 16.24\ci{15.63}{16.87} \\
 & FIGS & 72.64\sd{0.23} & 10.67 & 0.774 & 0.817 & 11.18\ci{10.50}{11.84} \\
 & RRL & 69.99\sd{0.34} & 13.98 & 0.767 & 0.991 & 11.82\ci{11.11}{12.61} \\
 & CLASSY & 68.59\sd{0.49} & 14.14 & 0.654 & --$^\ddagger$ & 16.88\ci{16.33}{17.43} \\
 & RIPPER & 66.59\sd{0.01} & 15.98 & 0.654 & --$^\ddagger$ & 16.89\ci{16.42}{17.37} \\
 & DR-Net & 66.08\sd{0.66} & 15.00 & 0.628 & --$^\ddagger$ & 17.80\ci{17.16}{18.43} \\
 & RL-Net & 65.89\sd{0.43} & 15.93 & 0.692 & 1.769 & 14.68\ci{14.16}{15.24} \\
 & CORELS & 65.86\sd{0.36}$^\dagger$ & 16.62 & 0.613 & --$^\ddagger$ & 18.37\ci{17.83}{18.86} \\
 & NeuRules & 65.83\sd{0.27}$^\dagger$ & 15.52 & 0.739$^\dagger$ & 0.462$^\dagger$ & 12.87\ci{12.08}{13.71} \\
 & GradTree & 62.16\sd{0.42} & 17.43 & 0.698 & --$^\ddagger$ & 16.03\ci{15.11}{16.92} \\
\midrule
Standard & CatBoost & \underline{78.78}\sd{0.20} & \underline{3.31} & \underline{0.850}$^\S$ & \underline{0.267}$^\S$ & \underline{1.61}\ci{1.37}{1.87} \\
classifiers & LightGBM & 78.71\sd{0.11} & 3.90 & 0.841$^\S$ & 0.279$^\S$ & 3.71\ci{3.21}{4.24} \\
 & XGBoost & 78.67\sd{0.37} & 4.21 & 0.842$^\S$ & 0.271$^\S$ & 3.42\ci{2.82}{4.05} \\
 & Random forest & 77.80\sd{0.05} & 4.90 & 0.828$^\S$ & 0.325$^\S$ & 6.53\ci{5.74}{7.32} \\
 & EBM & 77.86\sd{0.13} & 4.76 & 0.844$^\S$ & 0.299$^\S$ & 3.29\ci{2.76}{3.82} \\
  & GRANDE & 76.17\sd{0.44}$^\dagger$ & 8.60 & 0.823 & 0.321 & 6.76\ci{5.84}{7.82} \\
\addlinespace[2pt]\hdashline\addlinespace[2pt]
 (\emph{interpretable}) & Logistic regression & 73.72\sd{0.09} & 8.86 & 0.817$^\S$ & 0.396$^\S$ & 7.55\ci{6.42}{8.76} \\
\bottomrule
\end{tabular}
\end{table*}

\begin{figure*}[t]
\centering
\begin{minipage}[t]{0.47\textwidth}
	\centering
	\includegraphics[width=0.86\linewidth]{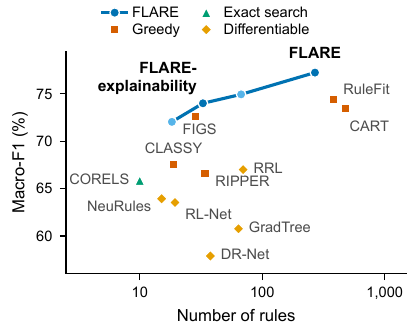}
	\caption{\textbf{Size of the rule base against performance on CC18.} The line shows FLARE with four sizes, from a single model with 20 rules to the final configuration. The light blue points are intermediate FLARE configurations. The same figure for TabArena is in Appendix~\ref{apx:capacity_category}.}
	\label{fig:size}
\end{minipage}\hfill
\begin{minipage}[t]{0.47\textwidth}
	\centering
	\includegraphics[width=0.86\linewidth]{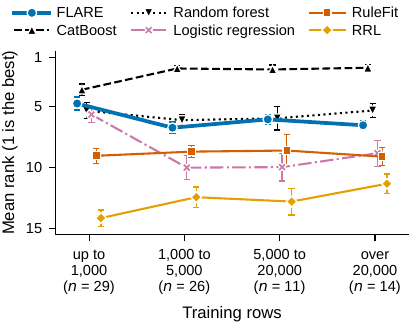}
	\caption{\textbf{Rank against the size of the dataset.} Mean rank of FLARE and of five of the other methods among the 20 methods of Table~\ref{tab:tabular}, on the 42 datasets of CC18 and the 38 tasks of TabArena together, grouped by their number of training rows.}
	\label{fig:scalability}
\end{minipage}
\end{figure*}

Table~\ref{tab:tabular} shows the results for the two tabular benchmarks. On CC18, FLARE obtains a macro-F1 of 77.67 and a mean rank of 5.29, which are better than those of all the rule and tree baselines. The second best rule learner is RRL, which is 7.69 points below, and does not win in any of the 42 datasets compared to FLARE. The rest of them are from 9.09 to 12.38 points below. The tree models are also significantly below, as RuleFit, CART and FIGS are from 3.27 to 5.03 points below FLARE, and even larger is the gap with GradTree (15.51 points below).
Regarding the standard classifiers, FLARE performs on par with random forests and EBM (statistical tests reported in Appendix~\ref{apx:tabular_paired}), it is 1.22 points above GRANDE and it is 3.95 points above logistic regression. 
However, the tuned gradient-boosted trees are still 1.00 to 1.11 points above FLARE, so there is still a gap with the best classifiers for tabular data.
The results in TabArena follow the same pattern. FLARE obtains a mean rank of 6.63, which is the best one among the rule and tree learners. Compared with the standard classifiers, FLARE is comparable to random forests and GRANDE. However, the difference with the gradient-boosted trees and EBM is larger than on CC18, with ROC AUC values of 0.841--0.850 against 0.825 for FLARE, and log-loss values of 0.267--0.299 against 0.348.

Figure~\ref{fig:size} shows how performance in relation with the size of the rule base for tree and rule based methods. For context, we show the performance of different FLARE operative points. FLARE-explainability has 33 rules on average, with 1.4 conditions each, and obtains a macro-F1 of 74.00. This is 3.24 points below FLARE with its default configuration (Figure~\ref{fig:size}), but it is still 4.02 points above RRL ($p<0.001$) and above the rest of the rule learners. It is also 1.36 points above FIGS, which has a similar size (28.5 rules), and it is 0.40 points below RuleFit, which has 378 rules ($p = 0.36$). On TabArena the picture is similar (Figure~\ref{fig:size_tabarena}): FLARE-explainability obtains a ROC AUC of 0.794 with 32 rules, which is above every rule learner except RuleFit (0.817 with 69 rules), and FLARE reaches 0.825, best overall in rule learning.

Finally, Figure~\ref{fig:scalability} shows the mean rank of FLARE and of some of the other methods on the datasets of the two benchmarks. FLARE obtains its best rank on the datasets with up to 1,000 rows (4.79), and from there it is between 6.09 and 6.77. In all the groups FLARE stays close to random forests and clearly ahead of the rule learners. Logistic regression goes from 5.69 to more than 8.8 beyond 1,000 rows.

\begin{table}[!htb]
\centering
\footnotesize
\setlength{\tabcolsep}{3pt}
\caption{\textbf{Differentiable rule learners with ablated components.} Macro-F1 reported. ($^*p<0.05$). Full results in Appendix: Tables~\ref{tab:apx_abl_published} and~\ref{tab:apx_abl_exchange}.}
\label{tab:study_published}
\begin{tabular}{@{}lccccc@{}}
\toprule
 & As & \multicolumn{2}{c}{Refitted weights} & \multicolumn{2}{c}{Other estimator} \\
\cmidrule(lr){3-4}\cmidrule(l){5-6}
Method & published & Own & Random & Best & Gain \\
\midrule
RRL & 67.00 & 71.34 & 66.26 & Soft & $-8.04^*$ \\
RL-Net & 63.53 & 66.64 & 52.49 & Grafting & $-5.11^*$ \\
DR-Net & 57.89 & 59.49 & 51.16 & Grafting & $+5.23^*$ \\
GradTree & 60.76 & 61.51 & 51.82 & ST-id & $+0.83$ \\
\midrule
Random trees & -- & -- & 72.24 & -- & -- \\
\bottomrule
\end{tabular}
\end{table}

\paragraph{What do differentiable rule learners need?}
\label{sec:results_study}
Table~\ref{tab:study_published} shows how the components of a rule learner affect different differentiable rule learners. We observed that refitting the weights of the rules once they have been found improves all of them. In fact, using 100 random rules in FLARE 1already beats RRL in Table~\ref{tab:tabular}. So, a large part of the performance of a rule learner comes from how its rules are weighted, even though search is still relevant according to our results in all methods but the DR-Net. Long rules are needed: a single FLARE model with 18 rules with 1 or 2 conditions is above every other differentiable rule learner, (72.04, Figure~\ref{fig:size}). Regarding how to relax the discrete search problem, it looks like every method prefers its own solution, except for the DR-Net. The full study and the ablation of FLARE components are in Appendix~\ref{apx:empirical_component_ablation} (Tables~\ref{tab:apx_abl_published}--\ref{tab:apx_abl_path}).

\subsection{Image and Graph Data}
\label{sec:results_other}

\begin{table}[t]
\centering
\footnotesize
\setlength{\tabcolsep}{3.5pt}
\caption{\textbf{Performance of the heads of a concept bottleneck model.} Macro-F1 (\%). The head is trained on the ground-truth concepts (GT) or on the predicted ones in a sequential training scheme (Pred.). DR-Net is evaluated only on Derm7pt due to computational constrains.}
\label{tab:image}
\begin{tabular}{@{}lcccccc@{}}
\toprule
 & \multicolumn{2}{c}{CUB} & \multicolumn{2}{c}{SUN} & \multicolumn{2}{c}{Derm7pt} \\
\cmidrule(lr){2-3} \cmidrule(lr){4-5} \cmidrule(l){6-7}
Head & GT & Pred. & GT & Pred. & GT & Pred. \\
\midrule
FLARE & 44.27 & \textbf{43.93}& \textbf{12.73} & \textbf{12.44} & 35.06 & \textbf{45.35} \\[-0.5ex]
 & \sd{1.27} & \sd{0.76} & \sd{0.64} & \sd{0.84} & \sd{2.06} & \sd{2.81} \\
\midrule
Logistic regression & \textbf{45.53} & 43.58 & 12.69 & 10.89 & \textbf{39.53} & 36.52 \\[-0.5ex]
 & \sd{0.72} & \sd{0.60} & \sd{0.82} & \sd{0.24} & \sd{4.21} & \sd{0.17} \\
\midrule
LogicCBM & 40.41 & 38.50 & 8.35 & 7.06 & 32.97 & 28.77 \\[-0.5ex]
 & \sd{1.02} & \sd{0.62} & \sd{1.01} & \sd{0.30} & \sd{1.15} & \sd{2.88} \\
RRL & 29.92 & 28.52 & 7.17 & 6.34 & 36.65 & 28.10 \\[-0.5ex]
 & \sd{0.54} & \sd{0.62} & \sd{0.11} & \sd{0.38} & \sd{2.17} & \sd{0.83} \\
DCR & 23.23 & 25.44 & 0.00 & 0.01 & 30.37 & 31.77 \\[-0.5ex]
 & \sd{2.43} & \sd{0.84} & \sd{0.00} & \sd{0.01} & \sd{3.03} & \sd{2.86} \\
NeuRules & 19.29 & 23.04 & 3.57 & 4.21 & 33.94 & 26.21 \\[-0.5ex]
 & \sd{0.26} & \sd{0.58} & \sd{0.20} & \sd{0.12} & \sd{1.90} & \sd{9.42} \\
GradTree & 3.64 & 2.81 & 0.21 & 0.17 & 32.64 & 30.44 \\[-0.5ex]
 & \sd{0.60} & \sd{0.33} & \sd{0.07} & \sd{0.07} & \sd{6.25} & \sd{5.72} \\
RL-Net & 3.28 & 3.84 & 1.11 & 1.00 & 34.01 & 26.25 \\[-0.5ex]
 & \sd{0.73} & \sd{0.93} & \sd{0.09} & \sd{0.27} & \sd{2.35} & \sd{3.92} \\
DR-Net & -- & -- & -- & -- & 13.07 & 10.45 \\[-0.5ex]
 &  &  &  &  & \sd{2.99} & \sd{2.94} \\
\bottomrule
\end{tabular}
\end{table}

Table~\ref{tab:image} shows the results of the different heads in concept bottleneck models. FLARE is the best rule-based head in five of the six settings. The exception is Derm7pt when the head is trained on the ground-truth concepts, where RRL is 1.59 points above it. In CUB and SUN the next rule-based head is LogicCBM, which is from 3.86 to 5.43 points below FLARE. We also note that several of the differentiable rule learners do not scale to hundreds of classes, as they obtain less than 4 points in CUB. Compared with logistic regression, which is the standard head, FLARE performs similarly in CUB and SUN. In Derm7pt, FLARE is 8.83 points above it when both are trained on the predicted concepts, but 4.47 points below when they are trained on the ground-truth ones.

\begin{table}[t]
\centering
\footnotesize
\setlength{\tabcolsep}{4pt}
\caption{\textbf{Results on graph data.} Area under the precision-recall curve (\%) in EMOGI and accuracy (\%) in the citation graphs. Mean and standard deviation over three seeds.}
\label{tab:graph}
\begin{tabular}{@{}lcccc@{}}
\toprule
Method & EMOGI & Cora & Citeseer & PubMed \\
\midrule
FLARE & 76.21 & 76.60 & 57.00 & 78.03 \\[-0.5ex]
 & \sd{0.80} & \sd{0.79} & \sd{1.56} & \sd{1.94} \\
\midrule
Logistic regression & 71.42 & 64.80 & 53.70 & 66.50 \\[-0.5ex]
 & \sd{0.00} & \sd{0.00} & \sd{0.00} & \sd{0.00} \\
SGC & 67.84 & 75.50 & 65.20 & 73.30 \\[-0.5ex]
 & \sd{0.32} & \sd{0.00} & \sd{0.00} & \sd{0.00} \\
GCN & 74.41 & 81.70 & 70.87 & 79.07 \\[-0.5ex]
 & \sd{0.40} & \sd{0.46} & \sd{0.38} & \sd{0.38} \\
GAT & -- & 83.27 & 71.40 & 77.77 \\[-0.5ex]
 &  & \sd{0.32} & \sd{0.26} & \sd{0.23} \\
Random forest & 77.03 & 75.27 & -- & 75.30 \\[-0.5ex]
 & \sd{0.11} & \sd{0.35} &  & \sd{0.17} \\
EBM & 78.50 & -- & -- & -- \\[-0.5ex]
 & \sd{0.48} &  &  &  \\
GRANDE & 77.66 & 73.77 & 61.87 & 79.10 \\[-0.5ex]
 & \sd{0.79} & \sd{1.31} & \sd{0.42} & \sd{0.30} \\
\midrule
RuleFit & 73.23 & 72.67 & 59.07 & 77.13 \\[-0.5ex]
 & \sd{0.36} & \sd{1.06} & \sd{0.93} & \sd{1.33} \\
RRL & 71.34 & 61.83 & 17.47 & 57.73 \\[-0.5ex]
 & \sd{0.90} & \sd{2.44} & \sd{1.27} & \sd{2.72} \\
CORELS & 41.14 & 42.70 & 30.80 & 64.70 \\[-0.5ex]
 & \sd{0.00} & \sd{0.00} & \sd{0.00} & \sd{0.00} \\
\bottomrule
\end{tabular}
\end{table}

Table~\ref{tab:graph} shows the results for graph data. FLARE's relative performance really varies in each task: it is close to the graph neural networks in PubMed, where it is 1.04 points below GCN and 0.26 above GAT, and in EMOGI it is above GCN and close to random forests and EBM. However, it is from 5.10 to 6.67 points below them in Cora, and the difference is much larger in Citeseer, which only has 120 labelled nodes for 3,703 variables. Among the rule learners, RuleFit is the closest to FLARE, as it is from 0.90 to 3.93 points below, and in Citeseer it is 2.07 points above. The rest of the rule and tree learners are below FLARE in every graph (Appendix~\ref{apx:graph_rule_learners}). %

\section{Conclusion and Discussion}
We started by profiling the performance of existing differentiable rule learners. We found that how the rules are weighted is the main factor driving performance in these methods. We also found that many short rules with a few interaction rules perform just as well or better than long rules, and the choice of gradient estimator depends on the learning method. Based on this study, we proposed FLARE, an interpretable rule-based classifier that combines a discrete search over short fuzzy rules with additive class weights. In our experiments, FLARE performed significantly better than any other rule learner on TabArena and CC18, and comparably to a random forest. As the head of a concept bottleneck model on image data, FLARE performed better than any other head over predicted concepts. On graph data, FLARE performed better than the rest of the rule learners in EMOGI, Cora and PubMed, but worse than graph neural networks in Cora and Citeseer. FLARE also scales well for large numbers of samples and features, with better performance-explainability trade-offs than the rest of the rule-based classifiers tested.
	
\paragraph{Limitations.} We have not measured human understanding or stability of the learned rules across seeds. The performance compared to the best gradient boosting baselines is not matched by any explainable method.  We only consider classification, and tasks of up to 100,000 training rows.

\ificmlshowauthors
\section*{Acknowledgements}
This research and Javier Fumanal-Idocin were supported by EU Horizon Europe under the Marie Skłodowska-Curie COFUND grant No 101081327 YUFE4Postdocs. Dr Javier Andreu-Perez is supported in part by UKRI BBSRC project EyeWarn (ref: APP37953). The authors acknowledge the use of the High Performance Computing Facility (Ceres) and its associated support services at the University of Essex in the completion of this work.
\fi

\bibliography{references}
\bibliographystyle{icml2026}

\newpage
\appendix
\onecolumn
\section{Notation}
\label{apx:notation}
A recap of the notation used throughout the paper, with explanations for each term, plus noting where each of them appeared for the first time.

\begin{center}
\small
\begin{tabular}{@{}p{0.2\textwidth}p{0.62\textwidth}l@{}}
\toprule
Symbol & Meaning & Defined in \\
\midrule
\multicolumn{3}{@{}l}{\emph{Model}} \\
$\mathbf{x}$, $x_j$ & A sample, and its value of variable $j$ & Sec.~\ref{sec:flare} \\
$R$, $r$ & Number of rules; a rule & Eq.~\eqref{eq:score} \\
$B$ & Number of models that are averaged & Sec.~\ref{sec:head} \\
$C$, $c$ & Number of classes; a class & Eq.~\eqref{eq:score} \\
$T_r(\mathbf{x})$ & Truth degree of rule $r$ for sample $\mathbf{x}$ & Sec.~\ref{sec:flare} \\
$s_c(\mathbf{x})$ & Score of class $c$ & Eq.~\eqref{eq:score} \\
$b_c$ & Bias of class $c$ & Eq.~\eqref{eq:score} \\
$w_{rc}$, $\mathbf{w}_r$ & Signed points of rule $r$ for class $c$; the vector of them, in $\mathbb{R}^C$ & Eq.~\eqref{eq:score}, Sec.~\ref{sec:head} \\
\midrule
\multicolumn{3}{@{}l}{\emph{Partitions of real variables}} \\
$L$, $l$ & Number of labels of a variable; a label & Sec.~\ref{sec:partitions} \\
$\mu_{jl}$ & Membership function of label $l$ of variable $j$ & Eq.~\eqref{eq:trapezoid} \\
$p_{j,i}$ & Breakpoints of the partition of $x_j$, $i = 0, \dots, 2L-1$ & Eq.~\eqref{eq:trapezoid} \\
$\delta_{j,i}$, $\boldsymbol{\delta}_j$ & Length of segment $i$ (plateau or ramp) of the partition of $x_j$; the vector of them & Sec.~\ref{sec:partitions} \\
$m_j$, $M_j$ & Smallest and largest training values of $x_j$ & Sec.~\ref{sec:partitions} \\
$\boldsymbol{\theta}_j$ & Learned logits of the segment lengths of $x_j$, in $\mathbb{R}^{2L-1}$ & Sec.~\ref{sec:partitions} \\
\midrule
\multicolumn{3}{@{}l}{\emph{Category spaces}} \\
$a$ & A category of a categorical variable & Sec.~\ref{sec:categorical} \\
$K$ & Dimension of the category space & Sec.~\ref{sec:categorical} \\
$\mathbf{z}_a$ & Point of category $a$, in $\mathbb{R}^K$ & Sec.~\ref{sec:categorical} \\
$\boldsymbol{\mu}_l$, $\sigma_l$ & Center and width of the Gaussian region of label $l$ & Sec.~\ref{sec:categorical} \\
$n_a$, $\pi_a$ & Training count of category $a$; its share of the training rows & Eq.~\eqref{eq:rec_loss} \\
$\mathbf{g}$, $g_a$ & Decoder; its probability for category $a$ & Eq.~\eqref{eq:rec_loss} \\
$\mathcal{L}_{\mathrm{rec}}$ & Reconstruction loss & Eq.~\eqref{eq:rec_loss} \\
$\mathcal{L}_{\mathrm{rad}}$ & Radius loss & Eq.~\eqref{eq:rad_loss} \\
$r_a$, $u_a$ & Target radius of category $a$; $u_a = n_a^{-1/2}$ & Eq.~\eqref{eq:target_radius} \\
$r_{\min}$, $r_{\max}$ & Range of the target radii & Eq.~\eqref{eq:target_radius} \\
\midrule
\multicolumn{3}{@{}l}{\emph{Rules}} \\
$S$, $s$ & Number of condition slots of a rule; a slot & Sec.~\ref{sec:rules} \\
$\alpha_{rsj}$ & Weight that slot $s$ of rule $r$ gives to variable $j$ & Eq.~\eqref{eq:selection} \\
$\beta_{rjl}$ & Weight that rule $r$ gives to label $l$ of variable $j$ & Eq.~\eqref{eq:selection} \\
$j_{rs}$, $l_{rj}$ & Variable selected by slot $s$ of rule $r$; label selected by rule $r$ for variable $j$ & Eq.~\eqref{eq:selection} \\
$\tau_{rs}(\mathbf{x})$ & Truth degree of the condition of slot $s$ of rule $r$ & Eq.~\eqref{eq:slot_truth} \\
$\gamma^{\mathrm{on}}_{rsj}$, $\gamma^{\mathrm{off}}_{rsj}$ & Weights of the switch of slot $s$ of rule $r$ for variable $j$ & Sec.~\ref{sec:rules} \\
$\mathcal{A}_r$ & Active slots of rule $r$ & Sec.~\ref{sec:rules} \\
\midrule
\multicolumn{3}{@{}l}{\emph{Training}} \\
$i$, $y_i$ & A row of the batch; its class & Eq.~\eqref{eq:ce_loss} \\
$p_c(\mathbf{x})$ & Probability of class $c$, the softmax of the scores & Eq.~\eqref{eq:ce_loss} \\
$\omega_c$ & Weight of class $c$ in the loss & Eq.~\eqref{eq:ce_loss} \\
$\mathcal{L}_{\mathrm{CE}}$ & Class-weighted cross-entropy & Eq.~\eqref{eq:ce_loss} \\
$\mathcal{L}$ & Training objective & Eq.~\eqref{eq:objective} \\
$\mathcal{J}$ & Variables with a category space & Eq.~\eqref{eq:objective} \\
$\lambda$ & Weight of the reconstruction loss & Eq.~\eqref{eq:objective} \\
$k$ & Uniform share of the backward surrogate in gradient grafting & Eq.~\eqref{eq:surrogate} \\
$Q$, $q$, $q^*$ & Number of candidates of a selector; a candidate; the chosen one & Eq.~\eqref{eq:surrogate} \\
$\phi_q$, $\rho_q$ & Weight of candidate $q$ in a selector; its softmax share & Eq.~\eqref{eq:surrogate} \\
$h_q$, $\tilde{h}$ & Truth of candidate $q$; the surrogate output of the selector in the backward pass & Eq.~\eqref{eq:surrogate} \\
$\eta$ & Temperature of the selectors' softmax & Eq.~\eqref{eq:surrogate} \\
$v_{rj}$ & Truth of the condition of rule $r$ on variable $j$ & Eq.~\eqref{eq:residual} \\
$\tilde{\ell}_r$ & Logarithm of the truth of rule $r$ with the residual term, used in the backward pass & Eq.~\eqref{eq:residual} \\
$T$ & Temperature of the probability recalibration & Sec.~\ref{sec:settings} \\
\bottomrule
\end{tabular}

\begin{tabular}{@{}p{0.2\textwidth}p{0.62\textwidth}l@{}}
\toprule
Symbol & Meaning & Defined in \\
\midrule
\multicolumn{3}{@{}l}{\emph{Differentiation (Appendix~\ref{apx:differentiation_specs})}} \\
$e_c$, $\mathbf e$ & Derivative of the cross-entropy with respect to the score of class $c$ & Eq.~\eqref{eq:apx_diff_score} \\
$d_r$ & Derivative of the loss with respect to $T_r$ & Eq.~\eqref{eq:apx_diff_head} \\
$\kappa_q$ & Coefficient of candidate $q$ in the surrogate of a selector & Eq.~\eqref{eq:apx_diff_kappa} \\
$V$ & Number of variables & Eq.~\eqref{eq:apx_diff_nesting} \\
$u_{rsj}$ & Truth of the condition of rule $r$ on variable $j$ after the switch of slot $s$ & Eq.~\eqref{eq:apx_diff_nesting} \\
$\tilde{\tau}_{rs}$, $\tilde{v}_{rj}$ & Surrogate outputs of the variable selector and of the label selector & Eq.~\eqref{eq:apx_diff_variable} \\
$t$ & Relative position of a value on a ramp & Eq.~\eqref{eq:apx_diff_ramp} \\
$\xi$ & Share of the epochs that remain & Eq.~\eqref{eq:apx_diff_schedule} \\
\bottomrule
\end{tabular}
\end{center}

\section{Empirical Study of the Components of Differentiable Rule Learners}
\label{apx:empirical_component_ablation}

A differentiable rule learner has to make several choices at once, such as which conditions form each rule, how the rules are weighted in the output layer, how many rules there are and how long they are, and how the gradient passes through the discrete choices. In this appendix we study these components separately, both for FLARE and for four published learners, in order to find which of them really affect performance. Our results point to four factors, which are the fitted weights of the output layer, a rule base with many one-condition rules plus a few interaction rules, bagging, and a forward pass that runs the same discrete rules that are deployed. Besides this, we find that condition search becomes significant when the input has thousands of variables, but it is not particularly effective in most already existing algorithms. FLARE is designed to combine all the results in this section to find the best possible rule learner.

\subsection{Protocol}

We use the same 42 datasets of OpenML-CC18 as in the main text, with a stratified 5-fold cross-validation and three seeds, and we report the macro-F1 on the test rows in percentage points. The results of each dataset are averaged over folds and seeds. In order to compare two configurations, we use a Wilcoxon signed-rank test over the 42 datasets, which are always evaluated on the same folds, and we report its $p$-value together with the number of datasets won, tied and lost (W/T/L) (we consider as a tie a difference of at most 0.5 points).

We study RRL \citep{wang2021}, RL-Net \citep{dierckx2023}, DR-Net \citep{qiao2021} and GradTree \citep{marton2024a}. The released version of DR-Net is binary, so we train one network per class in multi-class datasets. Each method is run in up to five different ways, which are described in Table~\ref{tab:apx_abl_protocol}.

\begin{table}[ht]
\centering
\small
\caption{\textbf{The five ways in which a method is run in this study.} The last two are only used for FLARE. Comparing the rules of a method with random rules, both with refitted weights, shows what the search of the structure found, while comparing the method as published with its own rules with refitted weights shows the effect of the output layer.}
\label{tab:apx_abl_protocol}
\begin{tabular}{@{}lp{0.76\textwidth}@{}}
\toprule
Name & What is run \\
\midrule
As published & The method as it is trained and deployed. \\
Refitted weights & The rules learned by the method (for a tree, its leaves) become 0/1 features of a class-balanced logistic regression, whose regularisation strength is chosen by cross-validation on the training rows. \\
Random structure & Use randomly-selected rules instead of the searched rules from the models. Rule class weights fitted as above. \\
No search & FLARE trained with no gradient to the selectors: the rules keep their random initial conditions and only the points and biases are learned. \\
Enumeration & FLARE with one rule for every pair of variable and label, so that no one-condition rule has to be searched for; only the points and biases are learned. \\
\bottomrule
\end{tabular}
\end{table}

We also test FLARE in the following configurations. The \emph{single model} has 100 one-condition rules, and the \emph{mixed rule base} adds 20 rules of up to two conditions to it. The \emph{bagged single model} merges eight single models, each of them trained with its own split of the training rows. Finally, \emph{FLARE-simple} is the mixed rule base with eight bags and with the slower step sizes of Section~\ref{sec:training} for the selectors and the interaction rules. FLARE then adds to it the ``at most'' and ``at least'' conditions, a third condition in the interaction rules and the learned crossings.

\subsection{The Fitted Weights of the Output Layer and the Searched Rules}

Table~\ref{tab:apx_abl_published} shows the results for the four published learners. If we keep the rules that each method learned and we only refit their weights with a logistic regression, all methods improve\rev{:} RRL by 4.35 points (W/T/L 38/3/1, $p<0.001$), RL-Net by 3.11 (31/11/0, $p<0.001$) and GradTree by 0.76 (18/23/1, $p<0.001$). For DR-Net the gain is 1.60 points, but it is not significant (13/21/8, $p=0.62$). So, in these methods, a final phase that only refines the class weights of the rules can have a significant impact.

These learners also find better rules than chance in the reversed experiment. With the rule class scores fitted with the same logistic regression, their own rules are from 5.08 to 14.15 points above random rules of the same size.

The table also includes a control baseline that does not search at all and is not matched to any learner here, which consists of 30 totally random trees of depth 5 (up to 960 leaves), with leaf decisions trained using the same logistic regression. This reference obtains a macro-F1 of 72.24, which is above the four learners tested (from 57.89 to 67.00), and also above their own rules with refitted weights (from 59.49 to 71.34). This means that a linear combination with real weights can obtain good results even with rules that nobody searched for, and that when using dense spaces of conditions, search becomes less relevant than simply fitting well the rules found.

\begin{table}[ht]
\centering
\small
\caption{\textbf{Published differentiable rule learners on CC18.} Macro-F1 on the 42 datasets with three seeds, for the ways of running a method of Table~\ref{tab:apx_abl_protocol}. ``Own $-$ random'' compares the two columns with refitted weights.}
\label{tab:apx_abl_published}
\begin{tabular}{@{}lcccccc@{}}
\toprule
 & As & Own rules, & Gain from & Random rules, & Own $-$ & W/T/L \\
Method & published & refitted weights & refitting & refitted weights & random & (own vs.\ random) \\
\midrule
RRL & 67.00 & 71.34 & $+4.35$ & 66.26 & $+5.08$ & 37/2/3 \\
RL-Net & 63.53 & 66.64 & $+3.11$ & 52.49 & $+14.15$ & 41/0/1 \\
DR-Net & 57.89 & 59.49 & $+1.60$ & 51.16 & $+8.34$ & 32/6/4 \\
GradTree & 60.76 & 61.51 & $+0.76$ & 51.82 & $+9.70$ & 39/1/2 \\
\midrule
Random trees & -- & -- & -- & 72.24 & -- & -- \\
\bottomrule
\end{tabular}
\end{table}

\subsection{Searched Rules Against Random Structure in FLARE}

For FLARE (Table~\ref{tab:apx_abl_structure}), the search adds 2.36 points to the single model and 2.45 to the mixed rule base with respect to the same models with their random initial rules. It also adds 0.92 points to the bagged single model and 1.77 to FLARE-simple, and the four differences are significant. The bagged single model is the one that gains the least, simply because the union of eight random rule bases covers many more conditions than a single one. On these datasets it is also possible to list every one-condition rule, and doing so is better than searching for 100 of them ($+0.59$, $p=0.02$). So, where enumeration is possible, the search does not improve on it, which makes sense, since search is only used because of the combinatorial explosion of rule search space). 

Rules of two conditions, however, cannot be enumerated. When we add them to the enumerated one-condition rules, 20 searched two-condition rules are better than 20 random ones ($+0.64$, $p=0.003$), but they only add 0.28 points over having none ($p=0.27$), while the random ones are worse than having none ($-0.37$, $p=0.02$). In the mixed rule base, the 20 two-condition rules add 0.44 points to the single model, which is not significant (16/15/11, $p=0.12$), and 0.77 points once both of them are bagged (19/16/7, $p=0.04$). This is the reason behind the average rule base of FLARE, in which most of the rules have a single condition and act as the main effects of an additive model, and only a small group is left for the interactions.

\begin{table}[ht]
\centering
\small
\caption{\textbf{FLARE against its controls on CC18.} Macro-F1 on the 42 datasets with three seeds. We show the mean of A and of B, their mean difference (computed before rounding), the datasets won, tied and lost by A, and the Wilcoxon $p$. In the last block, the 20 two-condition rules are added to the enumerated one-condition rules.}
\label{tab:apx_abl_structure}
\begin{tabular}{@{}lccccc@{}}
\toprule
Comparison (A against B) & A & B & A $-$ B & W/T/L & $p$ \\
\midrule
\multicolumn{6}{@{}l}{\emph{Searched rules against no search}} \\
Single model & 73.62 & 71.26 & $+2.36$ & 24/9/9 & $<0.001$ \\
Mixed rule base & 74.06 & 71.61 & $+2.45$ & 27/8/7 & $<0.001$ \\
Bagged single model & 75.70 & 74.79 & $+0.92$ & 19/15/8 & 0.03 \\
FLARE-simple & 76.75 & 74.98 & $+1.77$ & 27/11/4 & $<0.001$ \\
\midrule
\multicolumn{6}{@{}l}{\emph{Enumeration of every one-condition rule against the searched single model}} \\
Enumerated against searched & 74.21 & 73.62 & $+0.59$ & 23/10/9 & 0.02 \\
\midrule
\multicolumn{6}{@{}l}{\emph{Two-condition rules}} \\
Searched pairs against random pairs & 74.48 & 73.84 & $+0.64$ & 15/23/4 & 0.003 \\
Searched pairs against none & 74.48 & 74.21 & $+0.28$ & 15/17/10 & 0.27 \\
Random pairs against none & 73.84 & 74.21 & $-0.37$ & 7/23/12 & 0.02 \\
\bottomrule
\end{tabular}
\end{table}

How much the search adds depends on the number of candidate conditions with respect to the number of rules? In order to show this, we repeated the comparison outside tabular data, using the rule base of FLARE-simple and three seeds on fixed splits. As the head of a concept bottleneck model trained on annotated concepts, the eight FLARE bagged models hold 960 rules for about a hundred binary concepts. Here, the search does not really add anything: the macro-F1 is 39.07 with search and 39.46 without it on CUB (112 concepts, 200 classes), and 10.84 and 11.92 on SUN (102 concepts, 717 classes). In node classification, where the conditions describe a node and its neighbourhood, the search is also neutral on EMOGI, which has 256 variables (the AUPRC is 76.4 and 76.5 with search for the two neighbourhood encodings that we tried, and 77.0 and 76.6 without it). However, it is decisive on the citation graphs, whose variables are thousands of words. There, the accuracy goes from 45.0 and 51.7 to 70.6 and 72.6 on Cora (2,900 to 4,300 variables), from 70.4 to 78.0 on PubMed (3,500) and from 41.8 to 51.7 on Citeseer (11,109). The reason is that random conjunctions over thousands of sparse words almost never fire, so the output layer has nothing to combine.

Besides this, performance is not the only thing that the search provides. On four synthetic concept tasks, in which each class is a known conjunction over 10 concepts, every configuration of FLARE reaches a macro-F1 of 100 both with and without search. However, the share of those conjunctions that some rule computes exactly is 0.90 for FLARE-simple with search and 0.42 without it, and 0.46 and 0.08 for the single model.

\subsection{The Gradient Estimator}

\paragraph{Gradient estimation in the existing methods.} Each published learner comes with its own way of passing the gradient through its discrete choices. RRL uses gradient grafting, GradTree, DR-Net and RL-Net use a straight-through estimator, and NeuRules trains a soft model and reads the discrete one at the end. In order to see whether this choice is what limits them, in Table~\ref{tab:apx_abl_exchange} we train each learner with the estimators of the others, keeping everything else as published. The straight-through estimator replaces the derivative of each discrete operation by that of the soft version of it that the learner already has, and we also test the identity in its place and the clipped identity of DR-Net \citep{qiao2021}, which gives no gradient far below the threshold. Grafting computes both the discrete and the soft model, takes the loss at the discrete output, and passes the gradient through the soft one.

For three of the five learners (RRL, RL-Net and NeuRules), the estimator they were published with is the best, or not significantly worse than the best, on both benchmarks. Even the exchanges that help leave every learner more than ten points below FLARE on CC18. The only consistent gain observed is DR-Net with grafting, which goes from 57.98 to 63.21 on CC18  and from 60.79 to 62.08 on TabArena. GradTree also improves with the identity on TabArena, but not on CC18, so we draw no further conclusions from that. More importantly, several of this methods effectively stopped learning in several exchanges, making the learner predict a single class in most cases. This was the case inthe RRL with any of the straight-through estimators, because its activation has no gradient at the thresholded weights, and DR-Net and RL-Net with the identity.

\begin{table}[ht]
\centering
\small
\caption{\textbf{Published learners trained with the gradient estimators of the others.} Macro-F1 on the 42 CC18 datasets with three seeds, and on the 38 TabArena tasks with one seed (following TabArena own protocol). ST: straight-through with the learner's soft version of each operation; ST-id: with the identity; ST-clip: with the clipped identity of DR-Net. Bold is the native estimator of each learner, $^*$ marks a significant difference with it (Wilcoxon, $p<0.05$), and $=$ marks a cell that is the ST one by construction.}
\label{tab:apx_abl_exchange}
\begin{tabular}{@{}lcccccccccc@{}}
\toprule
 & \multicolumn{5}{c}{CC18} & \multicolumn{5}{c}{TabArena} \\
\cmidrule(lr){2-6}\cmidrule(l){7-11}
Learner & ST & ST-id & ST-clip & Grafting & Soft & ST & ST-id & ST-clip & Grafting & Soft \\
\midrule
RRL & 31.10$^*$ & = & 31.10$^*$ & \textbf{66.84} & 58.80$^*$ & 37.40$^*$ & = & 37.38$^*$ & \textbf{62.79} & 56.05$^*$ \\
RL-Net & \textbf{63.65} & 32.56$^*$ & = & 58.54$^*$ & 55.11$^*$ & \textbf{63.96} & 38.81$^*$ & = & 58.46$^*$ & 53.73$^*$ \\
DR-Net & \textbf{57.98} & 34.80$^*$ & = & 63.21$^*$ & 41.26$^*$ & \textbf{60.79} & 41.04$^*$ & = & 62.08$^*$ & 47.89$^*$ \\
GradTree & \textbf{60.71} & 61.54 & 54.85$^*$ & 59.59 & 57.07$^*$ & \textbf{60.23} & 63.62$^*$ & 52.92$^*$ & 58.33$^*$ & 56.41$^*$ \\
NeuRules & 59.55$^*$ & 52.62$^*$ & 53.02$^*$ & 62.90 & \textbf{63.96} & 60.29$^*$ & 53.84$^*$ & 49.11$^*$ & 60.31$^*$ & \textbf{63.70} \\
\bottomrule
\end{tabular}
\end{table}

\paragraph{Gradient estimation in FLARE.} In Table~\ref{tab:apx_abl_estimators} we change how the gradient reaches the selectors in FLARE. Gradient grafting lowers the uniform share $k$ of Eq.~\eqref{eq:surrogate} during training. The Gumbel softmax estimator \citep{jang2017} uses a soft of the candidates in the forward pass, according to their softmax weights, and anneals the temperature of that softmax from 1 to 0.01 during training. The identity and the clipped straight-through estimators send the gradient of the chosen candidate directly to the weights of all the candidates, the second one with the clip of DR-Net \citep{qiao2021}. Finally, the soft relaxation also uses the mixture of the candidates in the forward pass, and only reads the discrete rules at the end. Gradient grafting and the plain STE do not differ significantly, and neither does the identity STE. The Gumbel estimator is the only one above grafting, by 0.69 and 0.57 points on CC18 and by 0.71 and 0.51 on TabArena, and the four differences are significant. The clipped estimator, which gives no gradient to the candidates whose weight is far below that of the chosen one, loses from 1.65 to 4.17 points. %
The soft relaxation loses 8.43 and 7.80 points on CC18 and is worse on 40 and 39 of the 42 datasets, because the model that is trained is the most different from the one that is deployed.

\begin{table}[ht]
\centering
\small
\caption{\textbf{Gradient estimators for the selectors of FLARE.} Macro-F1 on the 42 CC18 datasets with three seeds, and on the 38 TabArena tasks with one seed (mean over the splits of each task), with everything else fixed. The differences and the Wilcoxon $p$ are computed against gradient grafting as baseline.}
\label{tab:apx_abl_estimators}
\begin{tabular}{@{}lcccccc@{}}
\toprule
 & \multicolumn{3}{c}{Single model} & \multicolumn{3}{c}{Mixed rule base} \\
\cmidrule(lr){2-4}\cmidrule(l){5-7}
Estimator & F1 & Difference & $p$ & F1 & Difference & $p$ \\
\midrule
\multicolumn{7}{@{}l}{\emph{CC18}} \\
Gradient grafting ($k$ lowered during training) & 73.60 & -- & -- & 74.17 & -- & -- \\
Straight-through ($k$ fixed at 1) & 73.02 & $-0.58$ & 0.54 & 74.16 & $-0.01$ & 0.96 \\
Straight-through, identity & 73.71 & $+0.11$ & 0.78 & 74.29 & $+0.12$ & 0.85 \\
Straight-through, clipped & 69.78 & $-3.82$ & $<0.001$ & 69.99 & $-4.17$ & $<0.001$ \\
Gumbel softmax & 74.29 & $+0.69$ & 0.02 & 74.74 & $+0.57$ & 0.01 \\
No gradient to the structure & 71.24 & $-2.36$ & $<0.001$ & 71.57 & $-2.60$ & $<0.001$ \\
Soft relaxation & 65.17 & $-8.43$ & $<0.001$ & 66.37 & $-7.80$ & $<0.001$ \\
\midrule
\multicolumn{7}{@{}l}{\emph{TabArena}} \\
Gradient grafting ($k$ lowered during training) & 68.42 & -- & -- & 68.99 & -- & -- \\
Straight-through ($k$ fixed at 1) & 68.60 & $+0.18$ & 0.43 & 69.29 & $+0.30$ & 0.11 \\
Straight-through, identity & 68.67 & $+0.26$ & 0.14 & 69.20 & $+0.21$ & 0.15 \\
Straight-through, clipped & 66.77 & $-1.65$ & $<0.001$ & 67.11 & $-1.88$ & $<0.001$ \\
Gumbel softmax & 69.13 & $+0.71$ & $<0.001$ & 69.50 & $+0.51$ & 0.02 \\
No gradient to the structure & 66.16 & $-2.25$ & $<0.001$ & 66.56 & $-2.43$ & $<0.001$ \\
Soft relaxation & 62.19 & $-6.22$ & $<0.001$ & 63.06 & $-5.93$ & $<0.001$ \\
\bottomrule
\end{tabular}
\end{table}

\paragraph{Studying the effect of the Gumbel estimator.} The gain of the Gumbel estimator depends on the configuration of FLARE. With FLARE-explainability, a single model of 40 rules, it significantly raises the macro-F1 on CC18 from 74.00 to 75.09 (29/10/3 datasets, $p<0.001$) and lowers the error on 27 of the 38 TabArena tasks ($p=0.004$; the ROC AUC goes from 0.794 to 0.804). However, in the bagged version of FLARE, which averages eight bags of 120 rules, it does not produce any significant improvement. We believe this is interesting to understand how gradient flows through FLARE.

\begin{figure}[ht]
	\centering
	\includegraphics[width=5.6in]{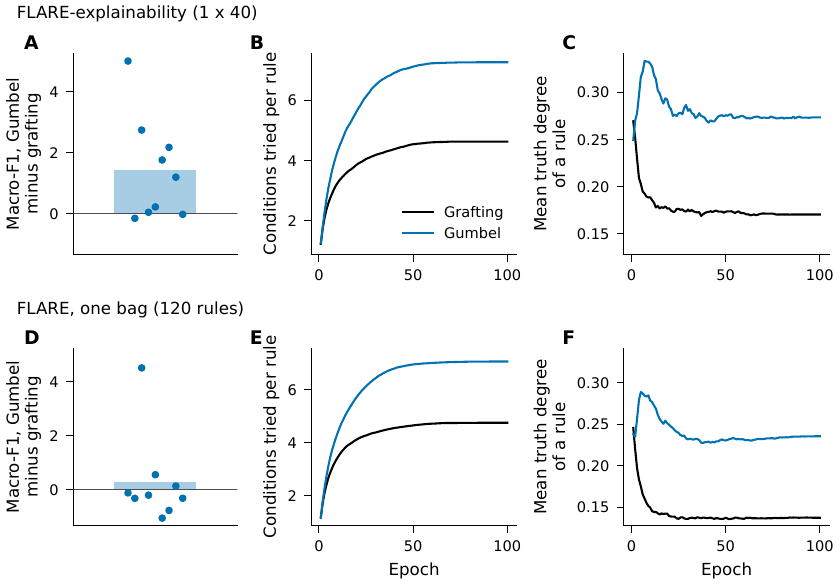}
	\caption{\textbf{Training with the Gumbel estimator and with gradient grafting, in FLARE-explainability (top) and in one submodel of FLARE (bottom).} A, D: macro-F1 of the Gumbel estimator minus that of grafting on the nine CC18 datasets tested. B, E: number of distinct conditions that each rule has held up to each epoch. C, F: mean truth degree of a rule over the training samples.}
	\label{fig:apx_gumbel}
\end{figure}

We followed the training of both configurations with each estimator on nine small CC18 datasets, two folds each, recording after every epoch the rules and their truth degree on each of the training samples. Figure~\ref{fig:apx_gumbel} shows two of these measures: the number of distinct conditions (variable and label) that each rule has held at some epoch, and the mean truth degree of a rule over the training samples. With grafting, the rules settle early on their first conditions: each rule tries 4.6 conditions over the whole training, and the rules become weaker as well. The noise of the Gumbel estimator keeps the choice of each rule open for longer: each rule tries 7.3 conditions (higher on the nine datasets, $p=0.004$), and their mean truth degree ends at 0.27 (higher on the nine datasets, $p=0.004$). In a control test, we removed the Gumbel noise and kept the soft forward pass and the lowered temperature. The rules are as strong as with the noise, but the search barely widens, and the macro-F1 does not improve. So, we can see here that for the single model Gumbel-Softmax widens the search on each rule, which is why results are improved.

Curiously, if we try this inside one bag of FLARE the Gumbel estimator, we obtain a similar result in terms of search (Figure~\ref{fig:apx_gumbel}, bottom row), as each rule now tries 7.1 conditions with Gumbel-Softmax instead of 4.8 with grafting. However, only the single model turns this into accuracy. We believe this is because FLARE averages eight different submodels, which already solves some of the search problems: a condition that one sub-model does not explore is likely to be explored by another.

\subsection{Summary of Contribution of the Components of FLARE}

Table~\ref{tab:apx_abl_path} expands the component study of \rev{Section~\ref{sec:results_study}} in the main text, adding to each step the datasets won, tied and lost and the Wilcoxon $p$. We can see there that adding the 20 two-condition rules is not significant on its own, while the other three steps are.

\begin{table}[ht]
\centering
\small
\caption{\textbf{Contribution of each component of FLARE on CC18.} Macro-F1 on the 42 datasets with three seeds. Each row adds one component to the row above, and the gain, the W/T/L and the Wilcoxon $p$ are computed against it.}
\label{tab:apx_abl_path}
\begin{tabular}{@{}p{0.52\textwidth}cccc@{}}
\toprule
Configuration & F1 & Gain & W/T/L & $p$ \\
\midrule
Single model with 100 searched one-condition rules & 73.62 & -- & -- & -- \\
$+$ 20 rules of up to two conditions (mixed rule base) & 74.06 & $+0.44$ & 16/15/11 & 0.12 \\
$+$ eight bags merged into one rule base & 76.48 & $+2.42$ & 36/6/0 & $<0.001$ \\
$+$ slower steps for selectors and interaction rules (FLARE-simple) & 76.77 & $+0.29$ & 13/25/4 & 0.03 \\
$+$ ``at most'' and ``at least'' conditions, up to three conditions in the interaction rules, learned crossings (FLARE) & 77.24 & $+0.49$ & 17/21/4 & $<0.001$ \\
\bottomrule
\end{tabular}
\end{table}

\subsection{What Did Not Help}

Here we also discuss other resources mentioned in the literature that did not work (as of the results obtained with them in CC18 benchmark), we comment on these, and the results obtained with them. We also discuss why they did not work and what conclusion we obtained from that experiment in the context of differentiable rule learning in general:

\begin{itemize}
	\item The Yager T-norm, which \citet{vankrieken2022} recommend for conjunctions. This did not work because it is exactly zero once the conditions of a rule are false enough, and no gradient returns from there. As a consequence, the rules of two conditions fall back to one (1.29 conditions per rule with the product and 1.09 with Yager), and about a third of the rules that decide some sample are lost (27.8 against 19.6). We stopped systematically testing for other T-norms after this result. 
	\item Sum only the largest contributions of each decision. In the literature, in a rule list the first rule that holds decides alone \citep{letham2015,angelino2018,dierckx2023,xu2025}, also, fuzzy classifiers have classically used only the winning rule, which is the one with the largest weighted truth degree \citep{fumanal2025compact}. In FLARE, this results in a small performance penalization when using the 20 largest values, and of one point on average in CC18 using only the 10 largest (mostly on multi-class datasets, $-2.44$ points there against $-0.32$ on the binary ones). For five rules only, we lost 3.17 points with the 5 largest firing strengths. So, the use of a limited number of rule s here definitively hurst performance, though the trade off might be worth for the user for example in binary classification problems.
	
	\item Letting each rule give points to a single class. This is the form of the rules in classical rule learning, where each rule concludes one class, as in RIPPER \citep{cohen1995ripper}, FURIA \citep{huhn2009furia}, decision sets \citep{lakkaraju2016} and genetic fuzzy classifiers \citep{alcalafdez2011fuzzy}. The rule lists learned by gradient descent also end with one class per rule \citep{dierckx2023,xu2025}.  However, results in FLARE were particularly conclusive. In the four datasets tried in CC18, the ones with more than two classes suffered performance degradation of different intensities, from 5.35 up to 27.61 points. The binary one did not change ($+0.64$), which is expected, because with two classes a signed point for one of them already counts against the other.
\end{itemize}

\section{Differentiation of FLARE}
\label{apx:differentiation_specs}

This appendix gives the derivatives used to train FLARE, in the order of backpropagation: from the loss to the rule truths, from a rule to its conditions, through the selectors of each condition and, finally, to the parameters of the memberships. We call the model ``discrete'' because the choices of variables, labels and switches are discrete, but note that their fuzzy truth degrees remain real-valued. In this sense, the forward pass is always the discrete model, in which every selector outputs the candidate it has chosen. The backward pass is ordinary backpropagation through the same computation, except that each selector is differentiated as its surrogate (Eq.~\eqref{eq:surrogate}) evaluated at the forward values. So, for a selector, $\partial\mathcal{L}/\partial(\cdot)$ denotes the signal given by gradient grafting, and not the derivative of the hard choice, which is zero almost everywhere.

The derivatives are written for one sample $\mathbf{x}$ of class $y$ in a batch, and the gradient of the cross-entropy with respect to a parameter is their sum over the rows of the batch. The losses of the category spaces are added separately, once per evaluation of the objective. We write $[\cdot]$ for the indicator that is 1 when the statement inside holds and 0 otherwise.

\paragraph{From the loss to the rule truths.}
Let $e_c$ be the derivative of the cross-entropy of Eq.~\eqref{eq:ce_loss} with respect to the score of class $c$. The score of Eq.~\eqref{eq:score} is linear in the biases, the points and the rule truths, so
\begin{gather}
	e_c = \frac{\partial \mathcal{L}_{\mathrm{CE}}}{\partial s_c} = \frac{\omega_{y}}{\sum_{i} \omega_{y_i}} \left(p_c - [c = y]\right),
	\label{eq:apx_diff_score} \\
	\frac{\partial \mathcal{L}}{\partial b_c} = e_c, \qquad
	\frac{\partial \mathcal{L}}{\partial w_{rc}} = e_c\, T_r, \qquad
	d_r = \frac{\partial \mathcal{L}}{\partial T_r} = \sum_{c=1}^{C} e_c\, w_{rc}.
	\label{eq:apx_diff_head}
\end{gather}
The derivative $e_c$ is the error of the predicted probability of class $c$, weighted by the class weight, and $d_r$ is proportional to the points that the rule gives on average under the current prediction minus the points it gives to the true class. So, when $d_r$ is negative, it means that the rule should favour the true class more than the current prediction does. The points of a rule learn from a sample only to the degree that the rule holds for it. Note that $d_r$ does not contain $T_r$, so the head still sends a signal to a rule even when the sample does not fire the rule. However, whether that signal reaches the parameters of the rule depends on the derivatives of the product and of the selectors that are given below. The class scores for each rule start from a normal distribution with standard deviation 0.1 and the biases from zero. Note that this does not guarantee a non-zero signal for every sample, and the signals of the samples of a batch can still cancel among themselves.

\paragraph{From a rule to its conditions.}
A slot that is switched off has truth 1 (Eq.~\eqref{eq:apx_diff_nesting} below), so the product of Eq.~\eqref{eq:rule_truth} is not affected by it. However, these slots still receive a derivative, through which they can be switched on again or move to another variable. Then,
\begin{equation}
	\frac{\partial \mathcal{L}}{\partial \tau_{rs}} = d_r \prod_{s' \neq s} \tau_{rs'}.
	\label{eq:apx_diff_slot}
\end{equation}
In words, a condition learns from a sample in proportion to how true the other conditions of its rule are. In a rule with a single active condition, the other slots have truth 1, so the condition receives $d_r$, whatever its own truth. In a rule with several conditions, each condition is weighted by the truths of the others. When exactly one condition is false, its sensitivity is $d_r$ times the truths of the other slots, while every other slot receives zero. So, a false condition can still learn while the rest of its rule holds, but a rule with two false conditions learns nothing from that sample. Note that Eq.~\eqref{eq:apx_diff_slot} is the sensitivity at the output of the slot, and the selectors below still scale it before it reaches their weights. If two conditions of a rule are false on every training sample, the rule receives no gradient from the cross-entropy at all, since its points, its selectors and its residual term (Eq.~\eqref{eq:apx_diff_condition}) are all multiplied by zero. FLARE restarts those rules randomly after each epoch.

We also tested the conjunction gradient of RRL \citep{wang2021} in the rules with several conditions, which differentiates $1/(1-\log T_r)$ instead of $T_r$. The derivative with respect to $\log T_r$ is then $1/(1-\log T_r)^2$ instead of $T_r$, and so it decays polynomially instead of exponentially in $\log T_r$. The effectiveness of this, however, is challenged by our empirical results. On the 42 CC18 datasets it changed the macro-F1 of a single model by $-0.13$ points on average (15 wins, 13 ties and 14 losses, Wilcoxon $p = 0.98$), so we kept the product as it was.

\paragraph{Derivatives of a selector.}
All the discrete choices of a rule (the variable of a slot, the switch of a slot and the label of a condition) are made by selectors of the form of Eq.~\eqref{eq:surrogate}. We first give the derivatives of a single selector, and then we apply them to the three selectors of a rule. Recall that a selector has the weights $\phi_1, \dots, \phi_Q$ over $Q$ candidates whose truths are $h_1, \dots, h_Q$, that the forward pass outputs the truth $h_{q^*}$ of the candidate with the largest weight, and that the backward pass differentiates the surrogate $\tilde{h}$ instead. Eq.~\eqref{eq:surrogate} can be written as a single sum,
\begin{equation}
	\tilde{h} = \sum_{q=1}^{Q} \kappa_q\, \rho_q\, h_q, \qquad \kappa_q = (1-k)\,[q = q^*] + \frac{k}{Q},
	\label{eq:apx_diff_kappa}
\end{equation}
where the coefficient $\kappa_q$ is $1 - k + k/Q$ for the chosen candidate and $k/Q$ for the others. The choice $q^*$ is not differentiated, so $\kappa_q$ is a constant in the backward pass, and $\tilde{h}$ depends on the weights only through the shares $\rho_q$.

The surrogate has two kinds of inputs, the truth values and the weights. The truths $h_q$ come from the computation below the selector (a membership, or the output of another selector), and their derivative is the part of the signal that the selector passes down. The weights $\phi_q$ decide the choice, and their derivative is what can change it. The first derivative follows directly from Eq.~\eqref{eq:apx_diff_kappa}. For the second one, the chain rule gives $\partial \tilde{h} / \partial \phi_q = \sum_{q'} \kappa_{q'} h_{q'}\, \partial \rho_{q'} / \partial \phi_q$, and the Jacobian of the softmax is $\partial \rho_{q'} / \partial \phi_q = \rho_{q'} \left([q' = q] - \rho_q\right) / \eta$. Hence
\begin{equation}
	\frac{\partial \tilde{h}}{\partial h_q} = \kappa_q\, \rho_q, \qquad
	\frac{\partial \tilde{h}}{\partial \phi_q} = \frac{\rho_q}{\eta} \left(\kappa_q h_q - \tilde{h}\right).
	\label{eq:apx_diff_selector}
\end{equation}
In the model, both derivatives are multiplied by the derivative $\partial \mathcal{L} / \partial \tilde{h}$ that the selector receives from above. Note the following properties of Eq.~\eqref{eq:apx_diff_selector} regarding selector choice:
\begin{itemize}
	\item \emph{How the choice changes.} When the loss asks for a larger truth ($\partial \mathcal{L} / \partial \tilde{h} < 0$), gradient descent raises the weights of the candidates whose weighted truth $\kappa_q h_q$ is above $\tilde{h}$ and lowers the others, so in this case selection does not change. However, when the reverse happens, the loss asks for a smaller truth, so it can happen that the choice in some selector changes.
	\item \emph{Gradient below the candidates that were not chosen.} A candidate that was not chosen has $\kappa_q = k/Q$, so it passes the gradient $(k/Q)\, \rho_q$ down to its truth. While $k > 0$, the gradient therefore also reaches the memberships and the selectors below the candidates that were not chosen. At $k = 0$ the weight of such a candidate still receives $-\rho_q \tilde{h} / \eta$ through the normalisation of the softmax, but its truth receives nothing.
\end{itemize}

We write $g = \partial \mathcal{L} / \partial \tilde{h}$ for the derivative that the selector receives from ``above'', and we compare the candidates that were not chosen through the derivative of their weights per unit of share, $(\partial \mathcal{L} / \partial \phi_q) / \rho_q$, since the shares $\rho_q$ differ between candidates.

\paragraph{The three selectors of a rule.}
Let $j = 1, \dots, V$ run over the variables, the real ones followed by the categorical ones. Slot $s$ of rule $r$ nests three selectors, each of the form of Eq.~\eqref{eq:surrogate}:
\begin{align}
	v_{rj} &= \mu_{j\, l_{rj}}(x_j) && \text{label selector, weights } \beta_{rjl} \text{, candidates } \mu_{jl}(x_j); \nonumber \\
	u_{rsj} &= \begin{cases} v_{rj} & \text{if } s = 1 \text{ or } \gamma^{\mathrm{on}}_{rsj} \ge \gamma^{\mathrm{off}}_{rsj}, \\ 1 & \text{otherwise} \end{cases} && \text{switch, weights } \gamma^{\mathrm{on}}_{rsj}, \gamma^{\mathrm{off}}_{rsj} \text{, candidates } v_{rj} \text{ and } 1; \label{eq:apx_diff_nesting} \\
	\tau_{rs} &= u_{rs\, j_{rs}} && \text{variable selector, weights } \alpha_{rsj} \text{, candidates } u_{rsj}. \nonumber
\end{align}
Here $v_{rj}$ is the truth of the condition of rule $r$ on variable $j$, and $u_{rsj}$ is that truth as slot $s$ sees it. For a real variable the label selector has $3L - 4$ candidates, which are the $L$ labels and the ``at most'' and ``at least'' sets, and so does a variable with a category space when these sets are used (for example, for ordered categories). The superscripts $\alpha$, $\beta$, on and off mark the shares $\rho$ and the coefficients $\kappa$ of the variable selector, of the label selector and of the two candidates of the switch, and $\tilde{\tau}_{rs}$ and $\tilde{v}_{rj}$ are the surrogate outputs of the first two. We apply Eq.~\eqref{eq:apx_diff_selector} from the outside in, starting from Eq.~\eqref{eq:apx_diff_slot}.

The variable selector decides which variable the slot uses, and its candidates are the truths $u_{rsj}$:
\begin{equation}
	\frac{\partial \mathcal{L}}{\partial \alpha_{rsj}} = \frac{\partial \mathcal{L}}{\partial \tau_{rs}}\, \frac{\rho^{\alpha}_{rsj}}{\eta} \left(\kappa^{\alpha}_{rsj}\, u_{rsj} - \tilde{\tau}_{rs}\right), \qquad
	\frac{\partial \mathcal{L}}{\partial u_{rsj}} = \frac{\partial \mathcal{L}}{\partial \tau_{rs}}\, \kappa^{\alpha}_{rsj}\, \rho^{\alpha}_{rsj}. \label{eq:apx_diff_variable}
\end{equation}
For $k<1$, in a slot, the selected variable receives the share $(1 - k + k/V)\, \rho^{\alpha}_{rsj}$ of the signal of the slot and every other variable $(k/V)\, \rho^{\alpha}_{rsj}$. While $k > 0$ the gradient therefore also reaches the label selectors and the partitions of variables that the rule does not use, which is how a rule can move to another variable. A slot that selects the variable of an earlier active slot of its rule is switched off in the forward pass ($\tau_{rs} = 1$, Section~\ref{sec:rules}), but the backward pass treats this replacement as an identity, so the slot keeps the derivatives above and can still move to another variable.

The switch decides whether the slot is used. Its candidates are the truth $v_{rj}$ of the condition and the constant 1 of a slot that is off, and the first slot has no switch, so $u_{r1j} = v_{rj}$. For $s \ge 2$,
\begin{equation}
	\frac{\partial \mathcal{L}}{\partial \gamma^{\mathrm{on}}_{rsj}} = -\frac{\partial \mathcal{L}}{\partial \gamma^{\mathrm{off}}_{rsj}} = \frac{\partial \mathcal{L}}{\partial u_{rsj}}\, \frac{\rho^{\mathrm{on}}_{rsj}\, \rho^{\mathrm{off}}_{rsj}}{\eta} \left(\kappa^{\mathrm{on}}_{rsj}\, v_{rj} - \kappa^{\mathrm{off}}_{rsj}\right). \label{eq:apx_diff_switch}
\end{equation}
The switch is a selector with $Q = 2$, and the following consequence of Eq.~\eqref{eq:apx_diff_switch} tells which samples can switch a slot on.

So, a condition is added to a rule pushed by the gradient from the samples that the rule should fire less less. This make sense if we realize that conjunctions are used to restrict the firing strength of a logical expression with additional conditions.

The condition of rule $r$ on variable $j$ is shared by all the slots of the rule, so its derivative collects the signals of all the slots that can read it (through the switch for $s \ge 2$), plus a residual term:
\begin{equation}
	\frac{\partial \mathcal{L}}{\partial v_{rj}} = \frac{\partial \mathcal{L}}{\partial u_{r1j}} + \sum_{s=2}^{S} \kappa^{\mathrm{on}}_{rsj}\, \rho^{\mathrm{on}}_{rsj}\, \frac{\partial \mathcal{L}}{\partial u_{rsj}} + \min(0.1, k)\, d_r\, T_r. \label{eq:apx_diff_condition}
\end{equation}
The last term of Eq.~\eqref{eq:apx_diff_condition} does not come from Eq.~\eqref{eq:surrogate}. It is the derivative of the residual of Eq.~\eqref{eq:residual}, which adds $\min(0.1, k) \sum_j v_{rj}$ to $\log T_r$ in the backward pass only, multiplied by $\partial T_r / \partial \log T_r = T_r$. This residual connection gives the condition of the rule on every variable a direct path to the rule truth, and it leaves the forward value unchanged. In the rules that are fixed to a single condition (Section~\ref{sec:rules}), the later slots stay switched off and the gradients of their $\alpha_{rsj}$, $\gamma^{\mathrm{on}}_{rsj}$ and $\gamma^{\mathrm{off}}_{rsj}$ are set to zero. However, their paths to the shared $v_{rj}$ are not removed, because a switch that is off still passes the gradient of the surrogate to its on candidate while $k>0$. So, fixing a rule to a single condition in the forward pass does not reduce its backward pass to that of a rule with a single slot.

Finally, the label selector decides which label the condition takes, and its candidates are the memberships of the labels of variable $j$, which all the rules share:
\begin{equation}
	\frac{\partial \mathcal{L}}{\partial \beta_{rjl}} = \frac{\partial \mathcal{L}}{\partial v_{rj}}\, \frac{\rho^{\beta}_{rjl}}{\eta} \left(\kappa^{\beta}_{rjl}\, \mu_{jl}(x_j) - \tilde{v}_{rj}\right), \qquad
	\frac{\partial \mathcal{L}}{\partial \mu_{jl}(x_j)} = \sum_{r=1}^{R} \kappa^{\beta}_{rjl}\, \rho^{\beta}_{rjl}\, \frac{\partial \mathcal{L}}{\partial v_{rj}}. \label{eq:apx_diff_label}
\end{equation}
The second derivative is a sum over the rules, and it is the signal that reaches the partitions and the category spaces. While $k > 0$, it also reaches the labels that no rule has chosen.

\paragraph{Partitions.}
For a real variable that is not constant, so that $M>m$ (we drop the index $j$) the breakpoints are $p_i = m + \sum_{i' \le i} \delta_{i'}$ with $\boldsymbol{\delta} = (M - m)\, \mathrm{softmax}(\boldsymbol{\theta})$ (Section~\ref{sec:partitions}), and $\boldsymbol{\theta}$ holds the only trainable parameters of the partition. From the Jacobian of the softmax,
\begin{equation}
	\frac{\partial p_i}{\partial \theta_{i'}} = \delta_{i'} \left( [i' \le i] - \frac{p_i - m}{M - m} \right), \qquad i, i' = 1, \dots, 2L-1,
	\label{eq:apx_diff_breakpoint}
\end{equation}
which is zero for the last breakpoint. A membership depends on the breakpoints only on a ramp. Let $x$ lie on the ramp shared by labels $l$ and $l+1$, $p_{2l-1} < x < p_{2l}$, at the relative position $t = (x - p_{2l-1}) / \delta_{2l}$. By Eq.~\eqref{eq:trapezoid}, $\mu_{l}(x) = 1 - t$ and $\mu_{l+1}(x) = t$. The ``at most'' and ``at least'' sets are sums of labels, so on this ramp ``at least $l+1$'' equals $t$, ``at most $l$'' equals $1 - t$, and all the other candidates of the label selector are constant (0 or 1). The loss thus depends on the partition only through $t$:
\begin{equation}
	\frac{\partial t}{\partial p_{2l-1}} = -\frac{1 - t}{\delta_{2l}}, \qquad
	\frac{\partial t}{\partial p_{2l}} = -\frac{t}{\delta_{2l}}, \qquad
	\frac{\partial \mathcal{L}}{\partial \theta_{i'}} = \frac{\partial \mathcal{L}}{\partial t} \left( \frac{\partial t}{\partial p_{2l-1}}\, \frac{\partial p_{2l-1}}{\partial \theta_{i'}} + \frac{\partial t}{\partial p_{2l}}\, \frac{\partial p_{2l}}{\partial \theta_{i'}} \right),
	\label{eq:apx_diff_ramp}
\end{equation}
where $\partial \mathcal{L} / \partial t$ is the sum of the derivatives of Eq.~\eqref{eq:apx_diff_label} for the candidates that equal $t$ minus the sum for those that equal $1 - t$. So, moving either end of the ramp moves the sample along it, and the loss decides through $\partial \mathcal{L} / \partial t$ whether the sample should belong more to label $l$ or to label $l+1$. When $x$ lies strictly inside a plateau, every candidate is constant and the sample gives the partition no gradient. Values outside $[m, M]$ are moved to the nearest end, so they lie on the first or the last plateau. Of the models that are averaged (Section~\ref{sec:head}), only the first one trains $\boldsymbol{\theta}$. The others read its partition as a constant, so that the averaged model has a single partition per variable.

\paragraph{Categorical variables.}
In the forward pass, $\mu_{jl}(x_j)$ is 1 for the category of the sample and 0 for the others. In the backward pass the other categories count as 0.1 (Section~\ref{sec:categorical}). So, $e_c$, $d_r$ and the derivatives of the points are computed with the forward truths, but Eqs.~\eqref{eq:apx_diff_slot}--\eqref{eq:apx_diff_label}, including the residual term, are evaluated with these softened memberships. Since the memberships of these variables are fixed, the backward pass ends at the weights $\beta_{rjl}$, with the derivative of Eq.~\eqref{eq:apx_diff_label}. Also, as in the backward pass such a condition is never exactly false, it does not cut the product of Eq.~\eqref{eq:apx_diff_slot}: on the samples of other categories, each categorical condition that does not match contributes a factor 0.1 to the sensitivities of the other slots, instead of 0. Several of them still multiply these factors, so the attenuation is reduced but not removed.

For a variable with a category space, the membership of category $a$ in label $l$ is $\mu_{jl}(a) = \exp\!\left(-\lVert \mathbf{z}_a - \boldsymbol{\mu}_l \rVert^2 / (2\sigma_l^2)\right)$, and
\begin{equation}
	\frac{\partial \mu_{jl}}{\partial \mathbf{z}_a} = -\frac{\partial \mu_{jl}}{\partial \boldsymbol{\mu}_l} = -\mu_{jl}\, \frac{\mathbf{z}_a - \boldsymbol{\mu}_l}{\sigma_l^2}, \qquad
	\frac{\partial \mu_{jl}}{\partial \log \sigma_l} = \mu_{jl}\, \frac{\lVert \mathbf{z}_a - \boldsymbol{\mu}_l \rVert^2}{\sigma_l^2},
	\label{eq:apx_diff_space}
\end{equation}
where the width is trained through its logarithm. So, a sample pulls a region towards its category, or pushes the region away from it, depending on the sign of the signal that the region receives, and it widens or narrows the region in the same way. When the ``at most'' and ``at least'' sets are used, they are the sums of the regions that they cover, capped at 1, so the signal that reaches a region is the sum of Eq.~\eqref{eq:apx_diff_label} over the candidates that contain it. Also, a Gaussian region has no flat interval, but its derivative with respect to the point is zero at the centre and can be very small far from it, and the contributions of several regions can cancel. The centres and the widths only receive the gradient of $\mathcal{L}_{\mathrm{CE}}$ that arrives through Eq.~\eqref{eq:apx_diff_label}. The points also receive the gradients of $\mathcal{L}_{\mathrm{rec}}$ and $\mathcal{L}_{\mathrm{rad}}$ in Eq.~\eqref{eq:objective}, and the decoder only that of $\mathcal{L}_{\mathrm{rec}}$. For completeness, we also show the derivatives of these two terms with respect to the points:
\begin{equation}
 \lambda\frac{\partial\mathcal L_{\mathrm{rec}}}{\partial\mathbf z_a}
 =-\lambda\pi_a\nabla_{\mathbf z_a}\log g_a(\mathbf z_a),\qquad
 \frac{\partial\mathcal L_{\mathrm{rad}}}{\partial\mathbf z_a}
 =2\pi_a(\lVert\mathbf z_a\rVert-r_a)\frac{\mathbf z_a}{\lVert\mathbf z_a\rVert},
 \label{eq:apx_diff_auxiliary}
\end{equation}
where the second one holds for $\mathbf z_a\ne0$. The first one moves each point to where the decoder recognises its category, and the second one moves it along its own direction towards its target radius. The decoder receives $-\lambda\sum_a\pi_a\nabla\log g_a(\mathbf z_a)$ with respect to its own parameters. 

\paragraph{Where does the gradient stops its flow?.}
There are several places where the gradient is very small or zero. We list them below, together with the parts of FLARE that tackles each problem.

\begin{itemize}
	\item \emph{Rules that do not hold.} The points of a rule receive a signal that is proportional to its truth, which is zero for a rule that does not hold (Eq.~\eqref{eq:apx_diff_head}). However, $d_r$ does not contain $T_r$, and in a rule with a single condition the condition receives $d_r$ whatever its own truth (Eq.~\eqref{eq:apx_diff_slot}), so such a rule can still move its condition until it holds for the samples. For multiple-condition rules, this does not work and we need to reset the rules after an epoch of lack of updates.
	\item \emph{Rules without a preference between the classes}. The conditions of a rule receive no gradient from a sample when $d_r = 0$, for example, when the rule gives the same points to every class. The random initialisation of the points avoids this at the start of training.
	\item \emph{Candidates that were not chosen.} For $k<1$ a candidate that was not chosen passes gradient to its truth only in proportion to $k$ (Eq.~\eqref{eq:apx_diff_selector}). For that reason, the schedule of $k$ starts at 1 and raises it again twice during training (Eq.~\eqref{eq:apx_diff_schedule}), the residual term gives the condition on every variable a path to the rule truth while $k > 0$, and the temperature of the selectors spreads the shares $\boldsymbol{\rho}$ over the candidates.
	\item \emph{Partitions.} A partition receives gradient only from the samples whose value lies on one of its ramps (Eq.~\eqref{eq:apx_diff_ramp}). The initial partition places the same share of the training rows on every plateau and ramp, and the softmax keeps every segment, and so every ramp, from collapsing to zero length (Section~\ref{sec:partitions}).
\end{itemize}

\section{Experimental Details}
\label{apx:experimental_details}

In this appendix we discuss the details that are needed to reproduce the experiments: the datasets, the preprocessing, all the hyperparameters of FLARE, and the implementation and the search space of each of the compared methods.

\subsection{Datasets}

Table~\ref{tab:apx_cc18} shows the 42 datasets of OpenML-CC18 used in the paper, and Table~\ref{tab:apx_tabarena} the 38 classification tasks of TabArena v0.1. In TabArena each task is evaluated with a 3-fold cross-validation, which is repeated 10 times for the smallest tasks and 3 times for the rest. This results in 30 or 9 train/test splits per task, and 594 in total. Eight of these tasks have at least one variable with more than 32 categories, with a maximum of 15,416. Finally, Table~\ref{tab:apx_other_data} shows the datasets used for concept bottleneck models and for graphs.

\begin{table}[ht]

\centering
\small
\caption{\textbf{The 42 datasets of OpenML-CC18 used in the paper.}}
\label{tab:apx_cc18}
\resizebox{\textwidth}{!}{\begin{tabular}{@{}lrrr@{\hskip 2em}lrrr@{}}
\toprule
Dataset & Rows & Variables & Classes & Dataset & Rows & Variables & Classes \\
\midrule
adult & 48{,}842 & 14 & 2 & ilpd & 583 & 10 & 2 \\
analcatdata\_authorship & 841 & 70 & 4 & jm1 & 10{,}885 & 21 & 2 \\
analcatdata\_dmft & 797 & 4 & 6 & jungle\_chess\_2pcs\_raw\_endgame\_complete & 44{,}819 & 6 & 3 \\
balance-scale & 625 & 4 & 3 & kc1 & 2{,}109 & 21 & 2 \\
bank-marketing & 45{,}211 & 16 & 2 & kc2 & 522 & 21 & 2 \\
banknote-authentication & 1{,}372 & 4 & 2 & kr-vs-kp & 3{,}196 & 36 & 2 \\
blood-transfusion-service-center & 748 & 4 & 2 & MiceProtein & 1{,}080 & 81 & 8 \\
breast-w & 699 & 9 & 2 & nomao & 34{,}465 & 118 & 2 \\
car & 1{,}728 & 6 & 4 & ozone-level-8hr & 2{,}534 & 72 & 2 \\
churn & 5{,}000 & 20 & 2 & pc1 & 1{,}109 & 21 & 2 \\
climate-model-simulation-crashes & 540 & 20 & 2 & pc3 & 1{,}563 & 37 & 2 \\
cmc & 1{,}473 & 9 & 3 & pc4 & 1{,}458 & 37 & 2 \\
connect-4 & 67{,}557 & 42 & 3 & PhishingWebsites & 11{,}055 & 30 & 2 \\
credit-approval & 690 & 15 & 2 & qsar-biodeg & 1{,}055 & 41 & 2 \\
credit-g & 1{,}000 & 20 & 2 & sick & 3{,}772 & 29 & 2 \\
cylinder-bands & 540 & 39 & 2 & steel-plates-fault & 1{,}941 & 27 & 7 \\
diabetes & 768 & 8 & 2 & tic-tac-toe & 958 & 9 & 2 \\
dresses-sales & 500 & 12 & 2 & vehicle & 846 & 18 & 4 \\
electricity & 45{,}312 & 8 & 2 & vowel & 990 & 12 & 11 \\
eucalyptus & 736 & 19 & 5 & wall-robot-navigation & 5{,}456 & 24 & 4 \\
first-order-theorem-proving & 6{,}118 & 51 & 6 & wdbc & 569 & 30 & 2 \\
\bottomrule
\end{tabular}
}
\end{table}

\begin{table}[ht]
\centering
\scriptsize
\caption{\textbf{The 38 classification tasks of TabArena.} For each task we show the training rows of one split, the number of variables (Var.), the number of classes (Cl.) and the number of train/test splits.}
\label{tab:apx_tabarena}
\resizebox{\textwidth}{!}{\begin{tabular}{@{}lrrrr@{\hskip 1.5em}lrrrr@{}}
\toprule
Task & Train.\ rows & Var. & Cl. & Splits & Task & Train.\ rows & Var. & Cl. & Splits \\
\midrule
Amazon\_employee\_access & 21{,}846 & 9 & 2 & 9 & hiva\_agnostic & 2{,}563 & 1{,}617 & 3 & 9 \\
anneal & 599 & 38 & 5 & 30 & HR\_Analytics\_Job\_Change\_of\_Data\_Scientists & 12{,}772 & 12 & 2 & 9 \\
APSFailure & 50{,}667 & 170 & 2 & 9 & in\_vehicle\_coupon\_recommendation & 8{,}456 & 24 & 2 & 9 \\
bank-marketing & 30{,}141 & 13 & 2 & 9 & Is-this-a-good-customer & 1{,}149 & 13 & 2 & 30 \\
Bank\_Customer\_Churn & 6{,}667 & 10 & 2 & 9 & jm1 & 7{,}257 & 21 & 2 & 9 \\
Bioresponse & 2{,}501 & 1{,}776 & 2 & 9 & kddcup09\_appetency & 33{,}333 & 212 & 2 & 9 \\
blood-transfusion-service-center & 499 & 4 & 2 & 30 & Marketing\_Campaign & 1{,}493 & 25 & 2 & 30 \\
churn & 3{,}333 & 19 & 2 & 9 & maternal\_health\_risk & 676 & 6 & 3 & 30 \\
coil2000\_insurance\_policies & 6{,}548 & 85 & 2 & 9 & MIC & 1{,}133 & 111 & 8 & 30 \\
credit-g & 667 & 20 & 2 & 30 & NATICUSdroid & 4{,}994 & 86 & 2 & 9 \\
credit\_card\_clients\_default & 20{,}000 & 23 & 2 & 9 & online\_shoppers\_intention & 8{,}220 & 17 & 2 & 9 \\
customer\_satisfaction\_in\_airline & 86{,}587 & 21 & 2 & 9 & polish\_companies\_bankruptcy & 3{,}940 & 64 & 2 & 9 \\
diabetes & 512 & 8 & 2 & 30 & qsar-biodeg & 703 & 41 & 2 & 30 \\
Diabetes130US & 47{,}679 & 47 & 2 & 9 & SDSS17 & 52{,}035 & 11 & 3 & 9 \\
E-CommereShippingData & 7{,}333 & 10 & 2 & 9 & seismic-bumps & 1{,}723 & 15 & 2 & 9 \\
Fitness\_Club & 1{,}000 & 6 & 2 & 30 & splice & 2{,}127 & 60 & 3 & 9 \\
GiveMeSomeCredit & 100{,}000 & 10 & 2 & 9 & students\_dropout\_and\_academic\_success & 2{,}949 & 36 & 3 & 9 \\
hazelnut-spread-contaminant-detection & 1{,}600 & 30 & 2 & 30 & taiwanese\_bankruptcy\_prediction & 4{,}546 & 94 & 2 & 9 \\
heloc & 6{,}973 & 23 & 2 & 9 & website\_phishing & 902 & 9 & 3 & 30 \\
\bottomrule
\end{tabular}
}
\end{table}

\begin{table}[ht]
\centering
\small
\caption{\textbf{Image datasets (top) and graph datasets (bottom).} The splits are given as training, validation and test samples.}
\label{tab:apx_other_data}
\begin{tabular}{@{}lrrll@{}}
\toprule
Dataset & Concepts & Classes & Split & Notes \\
\midrule
CUB, per image & 112 & 200 & 4,794 / 1,200 / 5,794 & annotations of each image \\
SUN Attribute & 102 & 717 & 11,472 / 1,434 / 1,434 & attribute votes cut at 0.5 \\
Derm7pt & 12 & 5 & 413 / 203 / 395 & split of \citet{kawahara2019seven} \\
Synthetic (4 tasks) & 10 & 5 & 2,000 / 500 / 1,000 & known conjunctions for each class \\
\midrule
Graph & Variables & Classes & Split & Nodes and metric \\
\midrule
Cora & 1,433 & 7 & 140 / 500 / 1,000 & 2,708 nodes, accuracy \\
Citeseer & 3,703 & 6 & 120 / 500 / 1,000 & 3,327 nodes, accuracy \\
PubMed & 500 & 3 & 60 / 500 / 1,000 & 19,717 nodes, accuracy \\
EMOGI & 64 & 2 & 746 test genes & 13,627 nodes, AUPRC \\
\bottomrule
\end{tabular}
\end{table}

\subsection{Preprocessing and Splits}

We fill the missing values with the median of the variable, or with its most frequent category if it is categorical, both computed on the training rows. Real variables are then standardised using the mean and the standard deviation of the training rows. FLARE reads the categorical variables directly, while the rest of the methods read them one-hot encoded. However, one-hot encoding is not feasible in TabArena for the variables that have thousands of categories. For that reason, for the baselines that we run there, a variable with more than 32 training categories is turned into a real one by replacing each category with its smoothed effect on the class, which is estimated on the training rows.

\subsection{Hyperparameters}

\paragraph{FLARE.}
Table~\ref{tab:apx_flare_hp} shows the values of the different configurations tested throughout the paper. Besides FLARE and FLARE-explainability, in the appendices we also use the other configurations of Table~\ref{tab:variants}, which we detail here as well. The changes that we make for concepts and graphs are in Appendix~\ref{apx:concepts_graphs}.

\begin{table}[ht]
\centering
\small
\caption{\textbf{Configurations of FLARE.} $R$ is given as the number of one-condition rules plus the number of interaction rules of each model. The pair limit only keeps, after training, the interaction rules of the 20 pairs of variables with the most evidence on the training rows.}
\label{tab:variants}
\begin{tabular}{@{}lcccl@{}}
\toprule
Configuration & $Bagged Models$ & $Rules$ & $S$ & Partitions \\
\midrule
FLARE & 8 & $100 + 20$ & 3 & learned \\
\quad pair limit & 8 & $100 + 20$ & 3 & learned \\
\quad one model & 1 & $100 + 20$ & 3 & learned \\
FLARE-explainability & 1 & $30 + 10$ & 3 & learned \\
\quad 20 rules & 1 & $15 + 5$ & 3 & learned \\
FLARE-simple & 8 & $100 + 20$ & 2 & fixed \\
\bottomrule
\end{tabular}
\end{table}

\begin{table}[ht]
\centering
\small
\caption{\textbf{Hyperparameters of FLARE for tabular data.}}
\label{tab:apx_flare_hp}
\begin{tabular}{@{}lll@{}}
\toprule
Hyperparameter & Value & Tuned in \\
\midrule
Labels per real variable & 7 & $\{5, 7, 9, 11\}$ \\
Rules per model & 120 & $\{60, 120, 240, 480\}$ \\
Share of one-condition rules & $100 / 120$ & $[0.6, 0.95]$ \\
Slots of an interaction rule & 3 & \\
Averaged models & 8 & $\{5, 8\}$ \\
Temperature of the selectors & 5 & $[2, 10]$ \\
Step of the points and biases & 0.1 & $[0.02, 0.3]$, log \\
Step of the selectors, relative to the points & 0.3 & $[0.1, 1]$, log \\
Factor of the interaction rules & 0.3 & $[0.1, 1]$, log \\
Step of the partitions, relative to the selectors & 0.01 & $[0.003, 0.03]$, log \\
Step of the category spaces, relative to the selectors & 0.1 & \\
Class weights in the loss & inverse frequency & on or off \\
Epochs & 100 & \\
Rows per batch & 32, or more (see text) & \\
Rows for validation & 20\% of the training rows & \\
Kept epoch & best validation accuracy & \\
Categories for a category space & more than 32 & \\
Dimension of a category space & 10 & \\
Hidden units of the decoder & 64 & \\
Weight of the reconstruction loss & 0.1 & \\
Range of the target radii & $[0.5, 3]$ & \\
Temperature of the recalibration (TabArena) & at least 1 & \\
\bottomrule
\end{tabular}
\end{table}

Regarding the training, each model is trained on 80\% of the training rows, and we keep the epoch in which the discrete model has the best accuracy on the remaining 20\%. The batch size grows with the dataset, so that an epoch never has more than 50 updates: we use 32 rows, or a fiftieth of the training rows when that is larger. The class scores for each rule are initialised from a normal distribution with a standard deviation of 0.1, and the biases at zero. The decoder of each category space has one hidden layer of 64 units.

The uniform share $k$ changes during the training. Let $\xi$ be the share of the epochs that remain when an epoch starts, so $\xi = 1$ in the first one. Then
\begin{equation}
	k = \xi + 0.2 \sin(4 \pi \xi),
	\label{eq:apx_diff_schedule}
\end{equation}
which is a linear decrease from 1 towards 0 with two oscillations that raise the uniform share again during training. With 100 epochs the last value is $k \approx 0.035$.

The parameters are updated by Adam \citep{kingma2015adam} in groups with different steps: 0.1 for the points and the biases, 0.3 times that step for the weights of the selectors ($\alpha_{rsj}$, $\beta_{rjl}$, $\gamma^{\mathrm{on}}_{rsj}$, $\gamma^{\mathrm{off}}_{rsj}$), 0.01 times the selectors' step for $\boldsymbol{\theta}$, and 0.1 times the selectors' step for the parameters of the category spaces. Every step decays along a cosine from its initial value to $10^{-6}$ over the epochs. In the rules with several conditions, each update of their selector weights and of their points is multiplied by 0.3 after the Adam step. Before each update, the gradient of all the parameters of a model is rescaled when its norm exceeds 1.

For tuned FLARE we keep the other hyperparameters fixed and we search the ranges in the last column of Table~\ref{tab:apx_flare_hp} for each fold, using the same budget as the baselines: 30 configurations proposed by the TPE sampler, or one hour. Each configuration is trained on 80\% of the training fold and evaluated with the macro-F1 on the remaining 20\%. The best configuration is then trained on the training fold in the same way as the default one.

\paragraph{Baselines.}
Table~\ref{tab:apx_baselines} shows the implementation used for each method and the hyperparameters that we tune. As the default configuration of a method we use the one in its package, or in the code released by its authors. Some of the methods had to be adapted so that they could be used in all the datasets. RIPPER, CORELS, DR-Net and RuleFit only support binary problems, so in multi-class datasets we train one model per class and assign each row to the class whose model fires. If none or several of them fire, we choose the most frequent class among the candidates. In TabArena, we report the results of its own methods when possible. 

\begin{table}[ht]
\centering
\scriptsize
\caption{\textbf{Implementation and search space of the compared methods.}}
\label{tab:apx_baselines}
\begin{tabular}{@{}lp{0.27\textwidth}p{0.58\textwidth}@{}}
\toprule
Method & Implementation & Tuned hyperparameters \\
\midrule
XGBoost & \texttt{xgboost}, histogram trees & trees $[100, 1000]$ log; depth $[2, 10]$; learning rate $[10^{-3}, 0.3]$ log; row and column subsampling $[0.5, 1]$; minimum child weight $[0.1, 10]$ log; L2 penalty $[10^{-3}, 10]$ log; class weights on or off \\
LightGBM & \texttt{lightgbm} & trees $[100, 1000]$ log; leaves $[8, 128]$ log; learning rate $[10^{-3}, 0.3]$ log; row and column subsampling $[0.5, 1]$; minimum rows per leaf $[5, 100]$ log; L2 penalty $[10^{-3}, 10]$ log; class weights on or off \\
CatBoost & \texttt{catboost} & trees $[100, 1000]$ log; depth $[3, 8]$; learning rate $[10^{-3}, 0.3]$ log; L2 penalty of the leaves $[1, 10]$ log; class weights on or off \\
RIPPER & \texttt{wittgenstein} & optimisation runs $[1, 3]$; share of pruning rows $[0.2, 0.5]$; description length allowance $[16, 128]$ log; bins $[5, 15]$ \\
CORELS & \texttt{corels}, 100,000 iterations & regularisation $[10^{-4}, 0.05]$ log; maximum conditions per rule $[1, 3]$; bins $[3, 9]$; minimum support $[0.005, 0.1]$ log \\
CLASSY & \texttt{rulelist} & maximum conditions per rule $[2, 5]$; beam width $\{25, 50, 100\}$; cut points $[3, 9]$; maximum rules $\{10, 20, 40\}$ \\
RRL & authors' code & width of the logical layer $\{32, 64, 128, 256\}$; learning rate $[5 \cdot 10^{-4}, 2 \cdot 10^{-2}]$ log; weight decay $[10^{-6}, 10^{-2}]$ log; binarisation nodes $[3, 10]$ \\
RL-Net & authors' code & rules $[10, 50]$ log; bins $[5, 15]$; learning rate $[10^{-3}, 5 \cdot 10^{-2}]$ log; sparsity of the conjunctions $[10^{-5}, 10^{-2}]$ log; L2 penalty $[10^{-6}, 10^{-2}]$ log \\
NeuRules & adapted from the authors' code & rules $[5, 30]$; learning rate $[10^{-3}, 5 \cdot 10^{-2}]$ log; minimum support $[0.05, 0.3]$; weight of the coverage term $[0.1, 1]$ \\
DR-Net & authors' code & rules $[20, 100]$ log; learning rate $[10^{-3}, 3 \cdot 10^{-2}]$ log; sparsity of the conjunctions $[10^{-4}, 10^{-1}]$ log; bins $[5, 15]$ \\
GradTree & our implementation in PyTorch, with the authors' defaults & depth $[3, 8]$; learning rates of the variable choices, the thresholds and the leaves, each $[10^{-3}, 0.1]$ log \\
GRANDE & \texttt{GRANDE} & depth $[3, 7]$; trees $\{256, 512, 1024\}$; learning rate of the weights $[10^{-4}, 0.25]$ log; learning rates of the variable choices, the thresholds and the leaves, each $[10^{-3}, 0.2]$ log; dropout $\{0, 0.25, 0.5\}$; share of variables per tree $[0.5, 1]$; class weights on or off \\
\midrule
Random forest & \texttt{scikit-learn}, 300 trees & variables per split $\{$square root, $\log_2$, 25\%, 50\%, 75\%, all$\}$; minimum rows per leaf $[1, 50]$ log; depth $[2, 32]$ log; class weights on or off \\
EBM & \texttt{interpret} & pairs of variables $[0, 20]$; learning rate $[5 \cdot 10^{-3}, 5 \cdot 10^{-2}]$ log; leaves per tree $[2, 4]$; minimum rows per leaf $[2, 20]$ log; smoothing rounds $\{0, 75, 200, 500\}$; class weights on or off \\
FIGS, RuleFit & \texttt{imodels}, at most 30 rules & maximum rules $[5, 50]$ log for FIGS and $[10, 100]$ log for RuleFit \\
CART & \texttt{scikit-learn} & depth $[2, 32]$ log; minimum rows per leaf $[1, 50]$ log; class weights on or off \\
Logistic reg. & \texttt{scikit-learn}, $C=1$ and no class weights by default & inverse of the L2 penalty $C$ $[10^{-3}, 10^{3}]$ log; class weights on or off \\
\bottomrule
\end{tabular}
\end{table}

\paragraph{Tuning.}
We explore different hyperparameter configurations, explored using TPE sampler of Optuna \citep{akiba2019optuna} with 30 configurations per fold or one hour, whichever comes first. Each configuration is fitted on 80\% of the training rows and evaluated with the macro-F1 on the remaining 20\%, and the best one is then fitted again on all of them. Training time is limited to 24 hours per method and dataset. All the methods use the same folds and the same preprocessing. 

\subsection{Rule Methods in Concept Bottleneck Models and Graphs}
\label{apx:concepts_graphs}

\paragraph{CBM in Image Datasets.} The concept detectors are Inception v3 networks pretrained on ImageNet and fine-tuned to predict the concepts for all methods. In FLARE, each concept $c \in [0,1]$ is treated as a variable with two fixed labels: \emph{true}, with truth $c$, and \emph{false}, with truth $1 - c$. There is no partition to learn in this case, so the rest is the same as in FLARE-simple.
Regarding the other heads, the Deep Concept Reasoner builds its rules from concept embeddings that are computed from the image. However, in our setting a head only reads the concepts, so we compute these embeddings from the vector of concepts instead. We run it with the settings of its paper (temperature 100), and also choosing the temperature among 100, 10 and 1 according to the macro-F1 on the validation samples. DR-Net, which needs one network per class, is not run in the datasets with more than 20 classes (it would require 200 models). The rest of the heads use the same default configurations as in the tabular experiments.

\paragraph{Graph node classification.} For graphs, each node is a sample that is described by its own variables. We also model in every node gets the share of its neighbours for which ``$x_j$ is $l$'' holds, computed with the truth degrees of the linguistic label. These shares are described with three linguistic labels (\emph{some}, \emph{many} and \emph{most}). Here we also start from FLARE-simple, but with $L = 3$ and with the same step size for the points, the selectors and the interaction rules, because with thousands of variables per node the rules need to move far from their initial conditions. Regarding the baselines, logistic regression is run on the variables of the node, and for the sake of fair comparison, we also add the means of its neighbours and the logarithm of its degree. Similarly, random forests and EBM use the variables of the node, the means of its neighbours at one or two hops and the logarithm of its degree, and SGC is run with one to three hops as well. In the citation graphs, GCN and GAT follow their published settings, with row-normalised variables.

\section{Expanded Results}
\label{apx:expanded_results}

In this appendix we give some additional results that complete Section~\ref{sec:results}.

\subsection{Paired Tabular Comparisons}
\label{apx:tabular_paired}

Table~\ref{tab:apx_tabular_paired} gives the paired comparisons that accompany the mean ranks of Table~\ref{tab:tabular} in both benchmarks. Each paired comparison uses the datasets or tasks completed by both methods, while the mean ranks of Table~\ref{tab:tabular} use all of them, assigning the rank 20 to the unfinished entries. In TabArena the test is computed on the error of each task, which is one minus the ROC AUC in the binary tasks and the log-loss in the multi-class ones.

\begin{table}[ht]
\centering
\small
\caption{\textbf{FLARE against each of the other methods.} On CC18 each method uses its tuned configuration when available, and on TabArena its default one. $n$ is the number of datasets or tasks that both methods finished, and $p$ is the two-sided Wilcoxon signed-rank test. On CC18, W/T/L counts the datasets won, tied and lost by FLARE, with a tie margin of 0.5 macro-F1 points. On TabArena, W/L counts the tasks in which FLARE has the lower and the higher error (one minus ROC AUC in the binary tasks, log-loss in the multi-class ones). $\Delta$ rank is the mean rank of the method minus the mean rank of FLARE in Table~\ref{tab:tabular}, so it is positive when FLARE ranks better and negative when the other method does.}
\label{tab:apx_tabular_paired}
\begin{tabular}{@{}lrrrrrrrr@{}}
\toprule
 & \multicolumn{4}{c}{CC18} & \multicolumn{4}{c}{TabArena} \\
\cmidrule(lr){2-5} \cmidrule(l){6-9}
Method & $n$ & W/T/L & $p$ & $\Delta$ rank & $n$ & W/L & $p$ & $\Delta$ rank \\
\midrule
\hspace{0.75em}-explainability & 42 & 38/4/0 & $<0.001$ & $+5.19$ & 38 & 36/2 & $<0.001$ & $+3.16$ \\
GRANDE & 36 & 17/13/6 & 0.007 & $+3.31$ & 37 & 18/19 & 0.521 & $+0.13$ \\
CART & 42 & 31/8/3 & $<0.001$ & $+5.45$ & 38 & 38/0 & $<0.001$ & $+9.61$ \\
FIGS & 42 & 36/3/3 & $<0.001$ & $+5.38$ & 38 & 38/0 & $<0.001$ & $+4.55$ \\
RuleFit & 42 & 33/5/4 & $<0.001$ & $+4.50$ & 38 & 26/12 & 0.003 & $+1.34$ \\
RRL & 42 & 41/1/0 & $<0.001$ & $+8.69$ & 38 & 38/0 & $<0.001$ & $+5.18$ \\
CLASSY & 42 & 41/1/0 & $<0.001$ & $+8.86$ & 38 & 38/0 & $<0.001$ & $+10.25$ \\
RIPPER & 42 & 42/0/0 & $<0.001$ & $+10.69$ & 38 & 38/0 & $<0.001$ & $+10.26$ \\
DR-Net & 42 & 40/2/0 & $<0.001$ & $+9.71$ & 38 & 38/0 & $<0.001$ & $+11.17$ \\
RL-Net & 42 & 42/0/0 & $<0.001$ & $+10.64$ & 38 & 38/0 & $<0.001$ & $+8.05$ \\
CORELS & 34 & 33/0/1 & $<0.001$ & $+11.33$ & 37 & 37/0 & $<0.001$ & $+11.74$ \\
NeuRules & 41 & 41/0/0 & $<0.001$ & $+10.24$ & 36 & 35/1 & $<0.001$ & $+6.24$ \\
GradTree & 42 & 41/1/0 & $<0.001$ & $+12.14$ & 38 & 38/0 & $<0.001$ & $+9.39$ \\
CatBoost & 42 & 6/11/25 & $<0.001$ & $-1.98$ & 38 & 0/38 & $<0.001$ & $-5.03$ \\
LightGBM & 42 & 9/12/21 & 0.009 & $-1.38$ & 38 & 3/35 & $<0.001$ & $-2.92$ \\
XGBoost & 42 & 10/8/24 & 0.016 & $-1.07$ & 38 & 4/34 & $<0.001$ & $-3.21$ \\
Random forest & 42 & 13/9/20 & 0.171 & $-0.38$ & 38 & 19/19 & 0.731 & $-0.11$ \\
EBM & 42 & 14/13/15 & 0.511 & $-0.52$ & 38 & 1/37 & $<0.001$ & $-3.34$ \\
Logistic regression & 42 & 30/5/7 & $<0.001$ & $+3.57$ & 38 & 21/17 & 0.144 & $+0.92$ \\
\bottomrule
\end{tabular}
\end{table}

\subsection{Example of a Rule Base}

Table~\ref{tab:apx_example_rules} shows the complete rule base of the model used in Figure~\ref{fig:overview}, which is FLARE-explainability trained on one of the folds of the adult dataset. The model has 35 rules, of which 27 have a single condition. Since the task is binary, for each rule we show the difference between the points that it gives to the two classes: positive points push the sample towards $\le$50K, and negative ones towards $>$50K. The score starts at $-1.13$, and each rule adds its points multiplied by its truth degree. The conditions on real variables use the labels of each partition, which are named after their position (from \emph{very low} to \emph{very high}), or with the names of Figure~\ref{fig:overview} in the case of age. A condition such as ``\emph{low or lower}'' is equivalent to the ``at most'' conditions of Section~\ref{sec:partitions}.

\begin{table}[ht]
\centering
\small
\caption{\textbf{Rule base of FLARE-explainability for one fold of the adult dataset.} The rules are sorted by the absolute value of their points. \textbf{Positive points push the sample towards $\le$50K and negative ones towards $>$50K}.}
\label{tab:apx_example_rules}
\begin{tabular}{@{}p{0.39\textwidth}r@{\hskip 1.5em}p{0.39\textwidth}r@{}}
\toprule
Rule & Points & Rule & Points \\
\midrule
capital-gain is very high & $-3.94$ & age is mature & $-0.59$ \\
capital-gain is high & $+3.42$ & fnlwgt is very low & $+0.58$ \\
relationship is Own-child & $+2.45$ & race is Black & $+0.57$ \\
relationship is Unmarried \textsc{and} sex is Female & $+2.36$ & education-num is fairly low or lower & $+0.50$ \\
age is very young \textsc{and} capital-gain is medium or lower & $+2.02$ & relationship is Unmarried & $+0.49$ \\
relationship is Not-in-family & $+1.79$ & age is young adult & $+0.47$ \\
relationship is Other-relative & $+1.59$ & sex is Female & $+0.33$ \\
capital-loss is high or higher & $-1.20$ & workclass is Self-emp-not-inc & $+0.25$ \\
hours-per-week is low or lower & $+1.16$ & capital-gain is low & $-0.23$ \\
age is young & $+1.13$ & age is very young & $+0.16$ \\
age is young or younger \textsc{and} marital-status is Never-married \textsc{and} hours-per-week is fairly high or lower & $+0.87$ & education is Prof-school & $-0.16$ \\
education-num is very low & $+0.84$ & capital-gain is fairly low & $-0.14$ \\
capital-loss is high & $-0.78$ & occupation is Tech-support & $-0.12$ \\
education-num is high or higher \textsc{and} marital-status is Married-civ-spouse \textsc{and} race is White & $-0.77$ & age is young adult or younger \textsc{and} native-country is Guatemala & $+0.08$ \\
occupation is Exec-managerial \textsc{and} hours-per-week is high or higher & $-0.76$ & workclass is Self-emp-not-inc \textsc{and} education-num is medium \textsc{and} marital-status is Never-married & $+0.07$ \\
education-num is very high & $-0.73$ & hours-per-week is medium & $-0.05$ \\
education-num is low \textsc{and} capital-gain is fairly high \textsc{and} native-country is Scotland & $-0.72$ & capital-gain is medium & $+0.02$ \\
relationship is Wife & $-0.64$ &  &  \\
\bottomrule
\end{tabular}

\end{table}

\subsection{Training Time}

Table~\ref{tab:apx_time} shows the time needed to train each method on one fold of the CC18 datasets with its default configuration. FLARE takes 35 seconds in the median dataset, and FLARE-explainability, which is a single model with fewer rules, takes 7 seconds. It should also be noted that these times are not directly comparable, because the methods that need a GPU use it and the rest of them run on a single CPU thread (Appendix~\ref{apx:experimental_details}). However, this also shows that in general, FLARE has a competitive running time, especially compared to other rule learners. 

\begin{table}[ht]
\centering
\small
\caption{\textbf{Training time in seconds for one fold of the CC18 datasets.} Median and mean over all the folds, with the default configuration of each method. $^\dagger$Over the runs that finished within the time limit.}
\label{tab:apx_time}
\begin{tabular}{@{}lrr@{\hskip 3em}lrr@{}}
\toprule
Method & Median & Mean & Method & Median & Mean \\
\midrule
FLARE & 34.5 & 121.8 & RRL & 8.3 & 87.9 \\
FLARE-explainability & 7.3 & 11.6 & RL-Net & 333.9 & 511.3 \\
XGBoost & 0.2 & 0.7 & NeuRules & 35.4 & 378.5 \\
LightGBM & 0.2 & 0.6 & DR-Net & 95.0 & 616.0 \\
CatBoost & 7.6 & 19.7 & GradTree & 4.4 & 17.7 \\
RIPPER$^\dagger$ & 2.2 & 66.2 & GRANDE$^\dagger$ & 672.6 & 2255.5 \\
CORELS & 1.4 & 32.6 & CLASSY & 16.9 & 58.1 \\
\bottomrule
\end{tabular}
\end{table}

\begin{figure}[ht]
	\centering
	\includegraphics[width=5.6in]{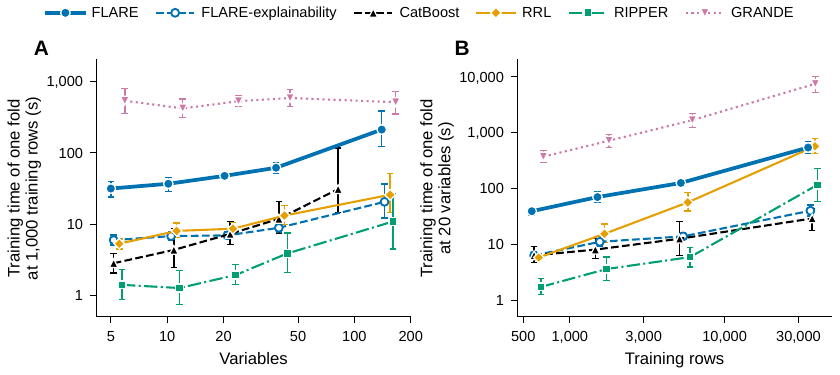}
	\caption{\textbf{Training time against the size of the dataset.} Geometric mean of the training time of one fold over the datasets of CC18 and the tasks of TabArena, grouped by their number of variables and by their number of training rows. The bars are the 95\% bootstrap interval of the mean over the datasets of the group. (CatBoost only uses the datasets of CC18, because its results on TabArena are the published ones).}
	\label{fig:time_variables}
\end{figure}

Figure~\ref{fig:time_variables} shows how the training time grows with the size of the dataset, taking the 42 datasets of CC18 and the 38 tasks of TabArena together. 
Since the datasets with more variables do not have the same number of rows, for each method we fit its time as a power of the training rows and a power of the variables, and we use these powers to show the time against the variables at 1,000 training rows, and against the training rows at 20 variables.

The time of FLARE grows with the number of variables to the power of 0.62 (95\% bootstrap interval over the datasets $[0.46, 0.71]$), and with the number of training rows to the power of 0.61 ($[0.56, 0.67]$). So, in both cases the growth is less than linear: a dataset with ten times more variables, or with ten times more rows, takes about four times longer. 
In the case of the variables, two of the tasks have more than 1,500 of them, and without these two the power is 0.53. For FLARE-explainability both powers are 0.43. Among the other methods, the power of the variables goes from zero for GRANDE, whose time does not depend on them, to 0.85 for CatBoost, and the power of the rows is close to one for RRL (1.11) and RIPPER (0.98), so RRL takes as long as FLARE on the datasets with more than 10,000 training rows. As in Table~\ref{tab:apx_time}, the height of the lines should not be compared between the methods that use a GPU and the ones that use a single CPU thread, only their slope.

\subsection{Size of the Rule Base}
\label{apx:capacity_category}

\begin{figure}[ht]
	\centering
	\includegraphics[width=5.6in]{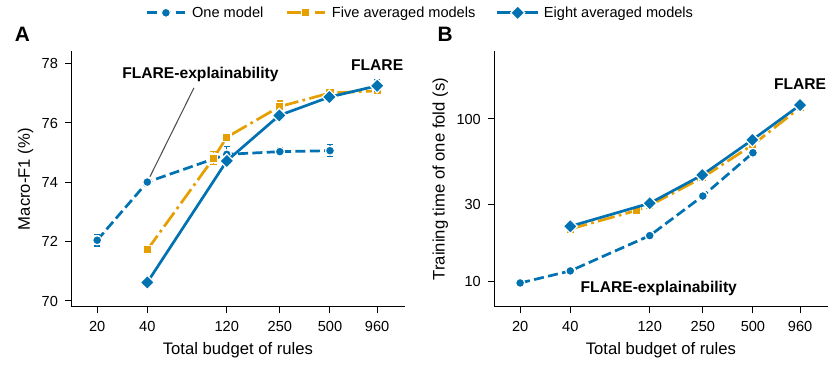}
	\caption{\textbf{FLARE with different sizes on CC18.} Macro-F1 (A) and training time of one fold (B) against the total budget of rules, which is the number of averaged models multiplied by the rules of each one: a single model with 20, 40, 120, 250 or 500 rules, five averaged models with 8, 20, 24, 50, 100 or 192 rules each, and eight averaged models with 5, 15, 31, 62 or 120 rules each, the last one being the FLARE configuration used in the main paper. Each point is the mean over the 42 datasets, and the bars are the standard deviation over three seeds. The number of different rules that each model keeps is in Table~\ref{tab:apx_stress}.}
	\label{fig:capacity}
\end{figure}

Figure~\ref{fig:capacity} and Table~\ref{tab:apx_stress} show how FLARE scales with its size, either with one larger model or by averaging five or eight models with the same total budget of rules, five sixths of them with a single condition, as in FLARE. A single model stops improving after 120 rules: it gains 1.96 points from 20 to 40 rules and 0.93 from 40 to 120, but only 0.12 more up to 500, where it needs 10.0 GiB of GPU memory. Averaging is worth once once each model has a few tens of rules. With 40 rules in total, the averages are 2.28 and 3.38 points below the single model, with 120 rules they are about the same, and from 250 rules on they are from 1.22 to 1.94 points above it, with little difference between five and eight models. Splitting the rules also keeps the memory cost down, since from 250 rules on an average takes at most 13 seconds more than the single model of the same budget and none needs more than 5.1 GiB in our experimentation.

\begin{table}[ht]
\centering
\small
\caption{\textbf{FLARE with different sizes on CC18.} Mean over the 42 datasets and seeds 33--35. The rules kept are the different rules of the trained model, the training time is that of one fold, and the GPU memory is the largest one over the datasets, measured on a GTX 1080 Ti with 11 GiB.}
\label{tab:apx_stress}
\begin{tabular}{@{}lrrrrr@{}}
\toprule
Models $\times$ rules & Budget & Macro-F1 & Rules kept & Time (s) & GPU memory (GiB) \\
\midrule
$1 \times 20$ & 20 & 72.04 & 18.3 & 9.8 & -- \\
$1 \times 40$ & 40 & 74.00 & 32.8 & 11.6 & -- \\
$1 \times 120$ & 120 & 74.94 & 67.2 & 19.2 & -- \\
$1 \times 250$ & 250 & 75.03 & 106.0 & 33.6 & 5.9 \\
$1 \times 500$ & 500 & 75.06 & 162.8 & 62.0 & 10.0 \\
\midrule
$5 \times 8$ & 40 & 71.72 & 26.3 & 21.1 & 2.6 \\
$5 \times 20$ & 100 & 74.81 & 56.4 & 27.4 & 3.1 \\
$5 \times 24$ & 120 & 75.51 & 66.1 & 29.4 & 3.4 \\
$5 \times 50$ & 250 & 76.55 & 110.2 & 43.7 & 3.2 \\
$5 \times 100$ & 500 & 77.00 & 177.1 & 69.5 & 4.9 \\
$5 \times 192$ & 960 & 77.07 & 269.6 & 117.6 & 5.1 \\
\midrule
$8 \times 5$ & 40 & 70.62 & 25.5 & 21.9 & 1.7 \\
$8 \times 15$ & 120 & 74.71 & 64.8 & 30.3 & 2.7 \\
$8 \times 31$ & 248 & 76.25 & 107.0 & 45.2 & 3.9 \\
$8 \times 62$ & 496 & 76.87 & 171.7 & 74.4 & 3.7 \\
$8 \times 120$ (FLARE) & 960 & 77.24 & 269.4 & 121.8 & 4.1 \\
\bottomrule
\end{tabular}
\end{table}

\begin{figure}[ht]
	\centering
	\includegraphics[width=5.6in]{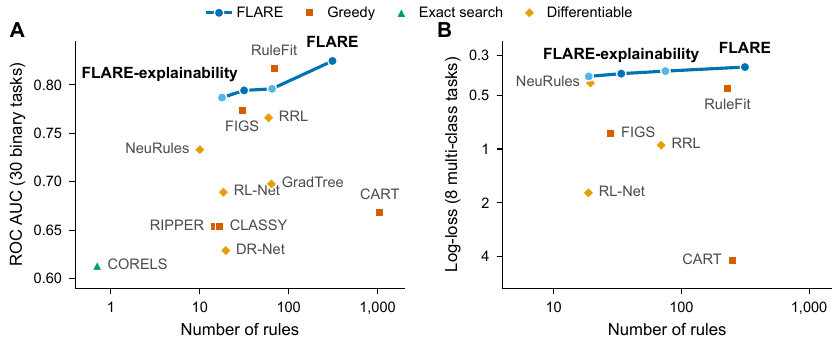}
	\caption{\textbf{Size of the rule base against performance on TabArena.} ROC AUC on the 30 binary tasks (A) and log-loss on the 8 multi-class tasks (B) against the number of rules of the same models. B has its axis is turned over, so that a point is better the higher it is in both panels. The line shows FLARE with four sizes, from a single model with 20 rules to the final configuration, and the light blue points are intermediate FLARE configurations.}
	\label{fig:size_tabarena}
\end{figure}

Figure~\ref{fig:size_tabarena} repeats Figure~\ref{fig:size} on TabArena, with the two scores of Table~\ref{tab:tabular} and the size of the same models: the ROC AUC on the 30 binary tasks, and the log-loss on the 8 multi-class ones. On TabArena all the methods use their default configuration and, following the protocol of the benchmark, each of them is evaluated once on each of the 9 or 30 outer splits of a task. The single models of FLARE obtain from 0.787 to 0.796 with 18 to 65 rules, and averaging eight of them reaches 0.825 with 311 rules. Among the other methods, only RuleFit is above the single models of FLARE, while FIGS, which has a similar size to FLARE-explainability, is 0.021 below it. The rule learners that only give a class, such as RIPPER, CLASSY and CORELS, stay below 0.66, since their ROC AUC is computed from the predicted class. In the multi-class tasks FLARE has the lowest log-loss with any of its sizes: the single models obtain from 0.393 to 0.367 with 19 to 75 rules, and the eight models 0.348 with 314 rules. NeuRules is the closest one, followed by RuleFit, while FIGS, RRL, RL-Net and CART are above 0.8.

\subsection{Rule and Tree Learners on Graph Data}
\label{apx:graph_rule_learners}

Table~\ref{tab:apx_graph_rule_learners} gives the results on graph data of the rule and tree learners that are not in Table~\ref{tab:graph}. As the ones in that table, they read the variables of each node and the means of its neighbours at one and two hops. All of them are below FLARE in every graph. In Cora and Citeseer, which only have 140 and 120 labelled nodes for thousands of variables, RL-Net and NeuRules predict almost only one class, and in the three citation graphs DR-Net keeps no rule and assigns every node to the first class, so its accuracy is the share of that class among the test nodes.

\begin{table}[ht]

\centering
\small
\caption{\textbf{Other rule and tree learners on graph data.} Area under the precision-recall curve (\%) in EMOGI and accuracy (\%) in the citation graphs, mean and standard deviation over three seeds. RIPPER, CLASSY, GradTree and DR-Net give no class probabilities, so their AUPRC is computed from the predicted class.}
\label{tab:apx_graph_rule_learners}
\begin{tabular}{@{}lcccc@{}}
\toprule
Method & EMOGI & Cora & Citeseer & PubMed \\
\midrule
FLARE & \textbf{76.21\sd{0.80}} & \textbf{76.60\sd{0.79}} & \textbf{57.00\sd{1.56}} & \textbf{78.03\sd{1.94}} \\
\midrule
FIGS & 65.38\sd{0.00} & 57.70\sd{1.23} & 46.83\sd{0.21} & 59.57\sd{1.10} \\
CART & 40.34\sd{0.15} & 65.13\sd{1.00} & 46.83\sd{2.10} & 59.97\sd{1.05} \\
CLASSY & 44.98\sd{0.00} & 47.70\sd{0.00} & 45.70\sd{0.00} & 49.90\sd{0.00} \\
RIPPER & 42.67\sd{1.85} & 49.07\sd{3.11} & 35.67\sd{1.72} & 44.27\sd{3.50} \\
RL-Net & 55.03\sd{1.83} & 16.13\sd{9.40} & 17.40\sd{1.21} & 41.97\sd{1.70} \\
GradTree & 42.05\sd{2.37} & 20.50\sd{1.84} & 19.43\sd{4.61} & 39.37\sd{7.44} \\
NeuRules & 56.67\sd{1.29} & 18.67\sd{1.29} & 7.83\sd{0.06} & 18.70\sd{0.26} \\
DR-Net & 50.95\sd{0.75} & 13.00\sd{0.00} & 7.70\sd{0.00} & 18.00\sd{0.00} \\
\bottomrule
\end{tabular}
\end{table}

\end{document}